# Stochastic Spiking Neuron Based SNN Can be Inherently Bayesian

Huannan Zheng[1], Jingli Liu[1], Kezhou Yang[1]*

[1] MICS Thrust, Function Hub, The Hong Kong University of Science and Technology (GuangZhou)

* kezhouyang@hkust-gz.edu.cn

**Abstract**

Uncertainty in biological neural systems appears to be computationally beneficial rather than detrimental. However, in neuromorphic computing systems, device variability often limits performance, including accuracy and efficiency. In this work, we propose a spiking Bayesian neural network (SBNN) framework that unifies the dynamic models of intrinsic device stochasticity (based on Magnetic Tunnel Junctions) and stochastic threshold neurons to leverage noise as a functional Bayesian resource. Experiments demonstrate that SBNN achieves high accuracy (99.16% on MNIST, 94.84% on CIFAR10) with 8-bit precision. Meanwhile rate estimation method provides a ~20-fold training speedup. Furthermore, SBNN exhibits superior robustness, showing a 67% accuracy improvement under synaptic weight noise and 12% under input noise compared to standard spiking neural networks. Crucially, hardware validation confirms that physical device implementation causes invisible accuracy and calibration loss compared to the algorithmic model. Converting device stochasticity into neuronal uncertainty offers a route to compact, energy-efficient neuromorphic computing under uncertainty.

**Introduction**

Over the past decade, deep learning–based artificial intelligence (AI) has developed rapidly (1). However, as tasks become more complex, the associated computational and energy costs grow sharply. In contrast, the human brain can perform similar tasks with much lower energy consumption (2). To approach this brain-level energy efficiency, neuromorphic computing paradigm seeks to construct highly efficient AI system by mimicking the functions of biological neural systems at both the algorithmic and hardware levels. Algorithmically, a critical mechanism is the emulation of biological information exchange via discrete spikes (3), which leads to the proposal of spiking neural networks (SNNs) computing model (4). Similar to biological brains, information is processed in a spiking manner in SNNs, which proves to reduce the power consumption in a variety of applications such as event camera automotive vision, edge computing and neuromorphic sensors (5)-(7). On the hardware side, neuromorphic computing paradigm aims to construct novel hardware systems where artificial neurons and synapses in the algorithm models can be directly emulated by the corresponding hardware units. In this way, the so-called von Neumann bottleneck problem due to the architecture mismatch between neural network and traditional computer can be eliminated, and the huge energy consumption in unnecessary data exchange can be saved. As an efficient way to construct hardware neurons and synapses, novel device technologies have been explored since they have underlying device physics similar to the behavior of biological neurons and synapses. Device technologies such as resistive random access memory (RRAM) (8)(9), phase-change memory (PCM) (10), and spintronic devices (magnetic tunnel junctions, MTJs) (11)(12) have shown great promise, and been demonstrated to construct device networks (13).

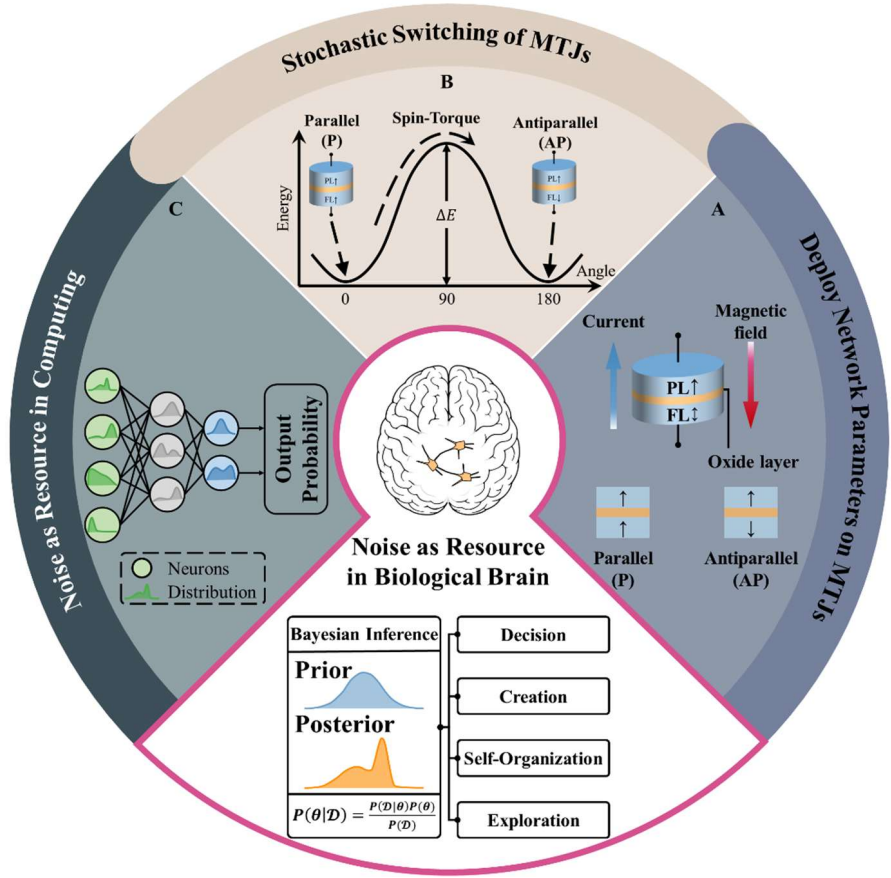

**Fig. 1. Bio-inspired Spiking Bayesian can be Inherently Bayesian.** Our proposed architecture is inspired by the utilization of neuronal noise in neural systems and is designed to utilize MTJ noise as a computational resource. Neural circuits achieve reliable computation and decision-making under neuronal uncertainty. The introduction of MTJ noise at the neuronal threshold is the underlying feature of neurons in our proposed architecture.

However, the performance of implementing algorithms onto hardware systems is still limited by device noise and low-precision parameters (14). When SNNs trained by algorithm are deployed on device systems, intrinsic device noise and stochasticity inevitably distort the target values (such as weight or neuron output values), which in turn degrades the accuracy and robustness of neuromorphic systems (15)-(17). As a result, the stochastic behavior is regarded as a negative factor to be eliminated rather than as a resource for computation (18)(19), and most engineering solutions aim to suppress such variability, for example by increasing barrier height or using error-correcting codes (20). These techniques improve reliability but also reduce the intrinsic energy efficiency of the devices.

On the other hand, similar stochastic dynamics are also widely observed in biological neural circuits, but such stochasticity appears to be computationally beneficial rather than detrimental (21)-(25). In biological brain, trial-to-trial variability supports Bayesian inference through the theory of neural sampling, where spike patterns are interpreted as samples from a posterior distribution, such as Gaussian or non-Gaussian distributions (26)-(29), to explicitly encode uncertainty (30)-(35). This sampling mechanism is the fundamental of important brain functions such as uncertainty representation (30)(33), sensory perception (26)(27), perceptual decision-making (21)(32)(34)(35), and confidence estimation(32)(35). This suggests that the intrinsic stochasticity of neuromorphic devices should be harnessed to emulate biological probabilistic computing, rather than suppressed (36)(37).

Based on the observation of stochastic behaviors in biological brains, hardware stochasticity is utilized to construct Bayesian neural network (BNN), which is a computing model proposed to conduct probabilistic inference (38)(39).The required stochasticity can be obtained from dedicated random-number generators (RNG) based on stochasticity in additional devices sources (40)-(42).  However, individual RNGs can be prohibitive in area and energy when scaled to large systems. Recently,  researchers showed that inherent stochasticity in synapse devices can be directly utilized for sampling required by BNNs (43)(44). Such method simplified the circuit design, but requires multi-bit synapse devices, which makes it difficult to encode precise weight values and scale down (43)(39). On the other hand, neuronal stochasticity, which is widely observed in biological circuits and represents various distributions, remains under-explored. Recent device studies have started to expose neuronal-level stochasticity at the hardware front-end. However, in most cases the stochasticity is either induced by external noise injection (45) or used only for encoding (46), rather than being intrinsically generated at the neuron event level and explicitly realized Bayesian inference as an in-network sampling mechanism akin to biological neuronal computation.

In our work, we proposed a Spiking Bayesian Neural Network (SBNN) framework that leverages the stochastic switching of binary magnetic tunnel junction (MTJ) devices as a neuronal uncertainty resource, as is shown in Fig. 1. We unify the dynamic models of stochastic MTJ devices and LIF neurons by mapping the stochastic switching behavior of MTJs to stochastic firing caused by random neuronal thresholds. In addition, we treat the threshold of stochastic neurons as a learnable parameter for optimization. This optimization mechanism acts as an additional long-term memory in the network (47), which enables SBNN outperforms full precision baselines on MNIST (achieving 99.16%) and FashionMNIST (achieving 90.05%) with only 8-bit precision weights and reaches a similar level on CIFAR10 (achieving 94.84%). In this framework, to reduce SNN training time, we introduce a rate estimation (RE) method that avoids the high cost of surrogate gradients (SG) and removes calibration related to conversion frameworks. Particularly, RE method provides roughly a 20-fold speedup at 64 time steps compared to SG method. On the other hand, with neuronal randomness, the membrane potential can exhibit multiple types of distributions at each time step. We compare membrane potential distributions induced by threshold randomness and by synaptic randomness in BNNs and find that threshold randomness better preserves the heavy tailed distributions common in neuroscience. Specifically, the Wasserstein-1 distance between membrane potential distribution and the normal Levy-stable distribution decreased with fan-in for weight uncertainty but remained comparatively elevated for neuronal uncertainty. This distribution property provides the network with a richer dynamic range and improved biological interpretability. During training, threshold optimization further equips SBNN with intrinsic Bayesian inference functionality. As a result, SBNN exhibits better calibration in accuracy and negative log likelihood (NLL) than existing BNNs across multiple network architectures. Moreover, this intrinsic uncertainty also provides stronger robustness than standard SNNs under noise injections into weights, neurons, and input images. Certainly, our SBNN get 67% accuracy improvement compared to standard SNN in weight uncertainty injection and 12% in input image noise attacks. Furthermore, this algorithm and hardware co-design not only benefits in accuracy or robustness but also naturally hardware friendly. Due to the unification of algorithm and device neuron model, the proposed neuron randomness aligns with device stochastic switching dynamics during training. As a result, this framework is suitable for deployment on the constraint hardware environment. Even with an extremely simple hardware structure, such as 784-10-10 with 8-bit weight precision, the model achieves more than 90% accuracy on MNIST.

Additionally, hardware experiments show that the device-level network attains invisible accuracy and NLL loss with the algorithmic model under the same parameters (MNIST, 784–20–10, binary MLP) across 100 images. Our SBNN is not limited to the MTJ device, and it can be transferred to other neuromorphic devices with stochastic physical characteristics. It therefore provides a general methodological basis for scalable, noise tolerant, and low latency brain inspired systems.

**Results**
**Stochastic Spiking Neuron Implemented with MTJ devices**
In this work, the stochastic spiking neuron is implemented using a magnetic tunnel junction (MTJ), the fundamental building block of spintronic devices (48). We established a mathematical relationship between the switching behavior of MTJ and the neuron dynamic of stochastic leaky integrate-and-fire (LIF) model, which making it possible for a device to work as a spiking neuron. As illustrated in Fig. 1(A), the MTJ comprises a pinned layer and a free layer separated by an oxide spacer, where the relative magnetic orientation determines the resistance state. When the magnetic moments of the two layers are in same (opposite) direction, the MTJ is in the parallel (anti-parallel) state and has a low (high) electrical resistance. Based on this correspondence, the parallel state can be associated with a spike event of the LIF spiking neuron model, and the anti-parallel state can be regarded as the silent state where no spikes fire. The standard LIF model is defined by,

$$\tau_m \frac{dV(t)}{dt} = -(V(t) - V_{reset}) + R_m I(t) \qquad (1)$$
$$S = H(V(t) - V_{th}) = \mathbf{1}[V(t) \geq V_{th}]$$

where $\tau_m$ is the membrane time constant, $I(t)$ is the input current, $V(t)$ is the membrane potential, and $V_{th}$ is the threshold. As shown in Fig. 1 (B), switching between parallel and anti-parallel states is driven by the spin-torque and is inherently stochastic due to thermal fluctuations near the critical voltage $U_C$. Consequently, the switching probability $P_{sw}$ is governed by the cumulative distribution function (CDF) of the threshold voltage,

$$P_{sw}(U(t)) = \int_{-\infty}^{U(t)} f_{MTJ}(v)\, dv = F_{MTJ}(U(t)) \qquad (2)$$

where $f_{MTJ}(\cdot)$ and $F_{MTJ}(\cdot)$ denote the probability density function (PDF) and CDF. In a deterministic LIF model, the firing probability $P_{det}$ is a step function, which can be written in the same integral form as Eq. (2),

$$P_{det}(V(t)) = \int_{-\infty}^{V(t)} \delta(v - V_{th})\, dv = H(V(t) - V_{th}) \qquad (3)$$

Here, $\delta(\cdot)$ is the Dirac delta function and $H(\cdot)$ is the Heaviside step function. Dirac delta function indicates the non-differentiable spiking event at the threshold $V_{th}$, while the step function represents the corresponding firing probability for a specific membrane potential $V(t)$. Under this view, deterministic firing corresponds to a determined value, whose CDF yields the hard step in Eq. (3). When introducing uncertainty to the threshold, its firing PDF changes from the Dirac function to a smooth PDF. The firing probability $P_{stoch}$ then evolves from the step function in Eq. (3) to a smooth CDF, consistent with the MTJ physics in Eq. (2),

$$P_{stoch}(V(t)) = \int_{-\infty}^{V(t)} f_{neu}(v)\, dv = F_{neu}(V(t)) \qquad (4)$$

where $f_{neu}(\cdot)$ and $F_{neu}(\cdot)$ are the PDF and CDF of the stochastic neuronal threshold. In this case, if we map the PDF of MTJ stochastic switching behavior, $f_{MTJ}(v)$, to the LIF threshold PDF, $f_{neu}(v)$, a single MTJ can be viewed as a stochastic LIF neuron. In other words, the stochastic critical voltage for MTJ switching can be viewed as the stochastic membrane threshold of stochastic LIF neurons.

On the other hand, in our hardware implementation, the MTJs are set to anti-parallel state at the beginning of each time step, waiting for input from previous layer(Table 1B), which can be viewed as a large effective leakage so that the membrane potential does not retain memory of previous states ($\tau_m \to 0$). In this limit, the LIF state reduces to an instantaneous mapping from the applied drive to $V(t)$. Using the applied MTJ voltage $U(t)$ as the input drive (with proportionality absorbed into $R_m$) and the stochastic threshold $\tilde{V}_{th}$ from Eq.(4), we obtain,

$$V(t) \approx R_m U(t), \qquad S(t) = H(V(t) - \tilde{V}_{th}) \qquad (5)$$

where $\tilde{V}_{th}$ follows the random distribution defined by $f_{MTJ}$. In this case, the input of stochastic LIF model corresponds to the applied voltage $U(t)$ of the MTJ, which found an explicit relation between the dynamic of stochastic LIF model and the behavior of MTJ. Table 1(C) shows that MTJs exhibit sigmoid probabilistic switching curves centered near 0.5 V, which indicates a Gaussian distribution for the threshold of membrane potential of the stochastic neurons, while Table 1(D) highlights their intrinsic thermal-noise-driven switching behavior.

**SBNN Architecture and Rate-Estimation Training**
**Hardware–Software Co-design for MTJ-based SBNN.** Based on the proposed MTJ-based stochastic neuron, we constructed a Spiking Bayesian Neural Network (SBNN) that realizes intrinsic threshold sampling, which works well even with low precision weights. In this Hardware-software co-design architecture, each neuron samples its threshold from a distribution defined by a learnable mean and variance. The learnable parameters are trained through the proposed rate-estimation training method. By optimizing these parameters through variational inference, the network naturally encodes Bayesian posterior distribution, enabling explainable probabilistic outputs. To validate architectural versatility, we incorporated these stochastic neurons into both fully connected networks (FCN) coupled by deterministic weights (Table 1J) and deep residual learning (ResNet) blocks (Table 1 K). This design exploits the device-level stochasticity to physically realize Bayesian inference, thereby minimizing the memory and circuit overhead typically associated with software-based random number generation.

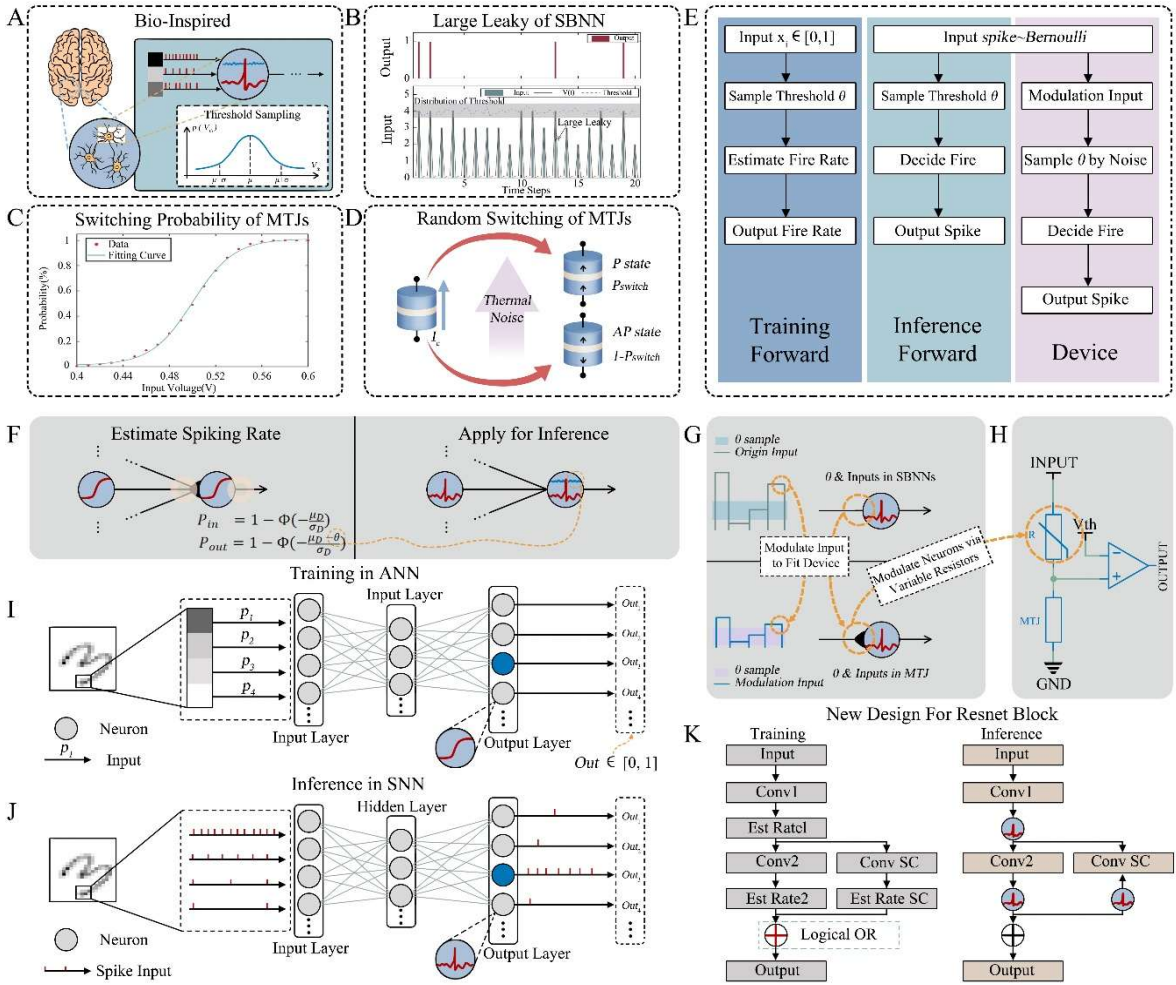

**Table 1. Spiking Bayesian Neural Network Implemented with MTJs.** (A) Bio-inspired Spiking Bayesian Neuron. Conceptual bio-inspired spiking Bayesian neuron (SBN) with brain and neuron sketches, and stochastic threshold sampling from a distribution. (B) Membrane Dynamics. Time-domain SBNN behavior showing that, identical inputs can produce different spike outputs. (C) MTJ Switching Probability. Experimentally measured probabilistic switching curve of a MTJ, showing a sigmoidal dependence of switching probability on the applied voltage, centered around 0.5 V. (D) Stochastic MTJ State Transitions. Identical current inputs may yield different resistance states with characteristic probabilities. (E) Training, Inference, and Device-Level Inference. The SBNN takes probability-valued inputs during training and binary spikes during inference. On-chip circuitry modulates the input. (F) Probabilistic SBN formulation. For a Gaussian pre-activation with mean $\mu_D$ and standard deviation $\sigma_D$, the input and output spike probabilities are $P_{\text{in}}$ and $P_{\text{out}}$, where $\theta$ is the stochastic threshold, and $\Phi$ is the standard normal cumulative-distribution function. (G) Input modulation. Algorithmic spike-count inputs are converted to scaled voltages or currents so that the effective MTJ input and threshold reproduce the desired switching probability (≈50% at the mean voltage in (C)), implemented via variable resistors. (H) Conceptual Hardware Neuron. Conceptual circuit in which a variable resistor $R$ (e.g., a multidomain MTJ) modulates the input to an MTJ-based stochastic neuron. (I) Training Forward probabilistic mode. Normalized image pixels $[0, 1]$ feed into sigmoid neurons, yielding probabilistic outputs during training. (J) Inference Forward Mode. The same parameters operate in an SBN-based spiking model. Inputs are Bernoulli-sampled spike events from pixel probabilities, producing binary spike outputs $\{0, 1\}$. (K) ResNet-compatible block design. During training,

branch fusion is performed using a logical-OR operation (red "+"), and during inference, it is replaced by linear residual addition (black "+").

**Rate-Estimation Training Method.** Conventional Spiking Neural Networks (SNNs) are constructed either by direct surrogate-gradient (SG) learning or by converting pretrained ANNs (49)(50)(51). However, SG methods rely on heuristic gradient approximations that often limit accuracy and convergence speed in deep architectures (52)(53), while conversion pipelines require extensive latency (hundreds of time steps) and post-hoc calibration to match firing rates (3)(54). To resolve these limitations, we developed a Rate-Estimation (RE) training framework (Table 1(I)) that enables efficient, end-to-end optimization by treating neural dynamics as the propagation of probability distributions rather than discrete events.

The pre-activation value of each time step, $D$, to a neuron is modeled as a sum of the independent Bernoulli trial from each neuron in previous layer multiplied by the corresponding weight value. By invoking the central limit theorem (CLT), the membrane potential distribution is characterized by the corresponding mean $\mu_D$ and variance $\sigma_D^2$, derived analytically from the input probability $P$ and synaptic weights $W$:

$$\mu_D = P \cdot W, \qquad \sigma_D^2 = (P \odot (1 - P)) * (W \odot W) \tag{6}$$

where $\odot$ denotes element-wise multiplication. However, directly calculating the activation, which is the output firing probability, $P_{out} = \Pr(D > \theta)$, suffers from high time complexity. We estimate this probability using a Probit-to-Logit approximation, which maps the Gaussian integral of the membrane potential to a scaled logistic sigmoid,

$$P_{out} = \sigma\left(k \cdot \frac{\mu_D - \theta}{\sigma_D + \varepsilon}\right) \tag{7}$$

Here, $\theta$ is the trainable stochastic threshold, $\varepsilon$ ensures numerical stability, and $k \approx 1.716$ scales the logistic curve to align with the normal PDF. This formulation circumvents the non-differentiable spike firing process, allowing gradients to be computed analytically during back-propagation. In this way, gradients are derived directly from the sigmoid-like PDF,

$$\frac{\partial P_{out}}{\partial \mu_D} \approx \frac{k P_{out}(1 - P_{out})}{\sigma_D + \varepsilon}, \qquad \frac{\partial P_{out}}{\partial \theta} = -\frac{\partial P_{out}}{\partial \mu_D} \tag{8}$$

These gradients are propagated to the weights $W$ via the chain rule applied to Eq. (6) (see Materials and Methods). By directly optimizing the probabilistic landscape, the RE method avoids the quantization errors of SG and the high latency of conversion methods. Consequently, the trained parameters in the algorithm faithfully reproduce the intrinsic stochasticity of the hardware and are transferred directly to the inference phase (Table 1(J)) without any parameter mapping or tuning required by ANN-to-SNN conversion method.

**Accuracy of SBNNs on Image Recognition Tasks**

To evaluate the performance of the SBNN trained with the proposed RE method, we conducted a series of supervised learning experiments on three widely used image classification benchmarks. We found that our SBNN achieved competitive performance among various datasets compared with other state-of-the-art models, as is shown in Table 1.

Moreover, we observed that the accuracy of our SBNN grew as the number of inference time steps increased and was affected little under the lower precision limitation. In each experiment, the training set was split into training set and validation set, and the test set was only used to test the model with the best performance on the validation set. For each dataset setting, we varied the number of inference time steps $T \in 2,3,4,6,8,16$.

**Table 1. Performance of SBNN in Image Recognition Tasks.** We tested our SBNN on different datasets. On MNIST, our SBNN performance exceeded that of general SNN and Binary NN. On Fashion MNIST, our performance (16 time steps) was almost outperforming to that of full-precision SNN (100 time steps). On Cifar10, our 8-bit mixed framework performance (8 time steps) was consistent with that of full-precision ANN2SNN (8 time steps), slightly exceeding that of Binary Neural Network.

| Method | Dataset | Architecture | Type | T | W/A (bits) | Accuracy (%) |
|---|---|---|---|---|---|---|
| BBB | | 784-800-10 | ANN | - | FP | 98.66 |
| | | 784-1200-10 | | | | 98.68 |
| Binary | | 784-1025(×3Hidden)-10 | | | 1/1 | 98.30 |
| Fast SNN | | | | 100 | FP | 97.91 |
| | MNIST | | | 2 | | 97.89 |
| | | | | 4 | 1/1 | 98.32 |
| Ours | | **784-1000-10** | SNN | 16 | | 98.71 |
| | | | | 2 | | 98.80 |
| | | | | 4 | **8/1** | 98.99 |
| | | | | 16 | | **99.16** |
| Binary | | 784-1024(×2Hidden)-10 | ANN | - | 1/1 | 85.00 |
| Fast SNN | | | | 100 | FP | 89.05 |
| | | | | 2 | | 83.72 |
| | Fashion | | | 4 | 1/1 | 85.78 |
| Ours | MNIST | **784-1000-10** | SNN | 16 | | 87.31 |
| | | | | 2 | | 85.92 |
| | | | | 4 | **8/1** | 88.38 |
| | | | | 16 | | **90.05** |
| BiPer | | Resnet-20 | ANN | - | 1/32 | 91.20 |
| QCFS | | Resnet-18 | | 8 | FP | 94.82 |
| IM-Loss | Cifar | | | 6 | FP | 95.40 |
| | 10 | | SNN | 2 | | 94.64 |
| Ours | | **Resnet-19** | | 4 | **Mixed†** | 94.61 |
| | | | | 8 | | **94.84** |

T denotes the time steps of SNN models.
W/A denotes the bit-width of weights (W) and activations (A).
**Mixed†**: In our implementation, ResNet-19 consists of two basic ResNet blocks and one EstRate block. The basic blocks use full-precision weights and activations, while the EstRate block and the final fully connected layer use 8-bit weights and spiking neurons with 1-bit activations.
**MNIST.** We evaluated SBNN on the MNIST digit-recognition benchmark, which contains 60,000 training and 10,000 test images of handwritten digits with a resolution of $28 \times 28$ pixels. We implemented a fully connected 784-1000-10 architecture under two precision regimes, including binary weights and activations (1/1) and 8-bit weights with binary activations (8/1). As shown in Table 2, the 8-bit SBNN demonstrated rapid convergence, achieving test accuracy of 98.80%, 98.99%, and 99.16% at latencies of $T = 2$, 4, and 16, respectively. Even with the fully binary configuration (1/1), the model reached 98.71% accuracy at $T = 16$. Despite reduced numerical precision and temporal depth, the SBNN outperformed full-precision baselines. The 8-bit model ($T = 16$) surpassed both the non-spiking Bayes-by-Backprop (BBB) (55) network (98.68%) and a deeper binary neural network (56) (98.30%). Notably, our approach exhibited superior efficiency, at just $T = 4$, the 8-bit SBNN exceeded the accuracy of a full-precision Fast SNN (57) (97.91%) that requires 100 time steps.

**Fashion MNIST.** We next evaluated SBNN on Fashion MNIST, which comprises 60,000 training and 10,000 test grayscale images of clothing items at 28 × 28 resolution. The architecture of our SBNN was an FCN (784-1000-10), the same as the reported Fast SNN. We report results at 2, 4, and 16 inference time steps.

As shown in Table 1, SBNN exhibited a clear accuracy gain with increasing time steps. With binary weights and activations, accuracy increases from 83.72% at $T = 2$ to 85.78% at $T = 4$ and 87.31% at $T = 16$. The $T = 4$ result exceeded the binary ANN baseline of 85.00% obtained with a deeper 784–1024×2–10 architecture (58). Increasing the weight precision to 8 bits further improved performance to 85.92% at $T = 2$, 88.38% at $T = 4$, and 90.05% at $T = 16$. At $T = 16$, this configuration surpassed the full-precision Fast SNN, which attained 89.05% with 100 inference steps. Together, these results indicated that SBNN can reach or even exceed established baselines while using low-precision parameters and substantially fewer time steps.

**Cifar10** We evaluated the performance of the proposed SBNN on CIFAR 10, which contains 60,000 color images of size 32 × 32 across 10 classes, with 50,000 training examples and 10,000 test examples [Cifar10 Ref]. We used ResNet-19, which is a similar backbone to the prior baselines. In our implementation, 8-bit quantization was applied only within the Spiking Bayesian blocks, while the remaining layers were kept in full precision.

As is shown in Table 1, our approach attained 94.64% accuracy at $T = 2$, 94.61% at $T = 4$, and 94.84% at $T = 8$. The $T = 8$ result was comparable to QCFS(59), which reported 94.82% at $T = 8$ using full precision weight values. Our result remained below IM Loss (60), which reached 95.40% at $T = 6$ with weight values in full precision. Compared to BiPer (61), which reported 91.20% on ResNet 20 with 1-bit weights and 32-bit activations, our method provided a substantial improvement while operating with a small number of inference steps.

Among these results in the different datasets, our SBNN framework performs competitive accuracy, which even surpasses the relative state-of-the-art (SOTA) models. We attributed this efficiency to the learned stochastic thresholds, which endow individual neurons with a memory-like function and enhance the expressive power of low-precision spiking dynamics, thereby narrowing the performance gap between low-precision SNNs and full-precision ANNs.

**Faster Training by Rate-Estimation Method**

To quantify the computational advantage of the proposed RE method, we compared its training dynamics with surrogate gradient training using the same SBNN model and an identical 784-1000-10 FCN architecture on MNIST. As is shown in Fig. 2 (A), changing time steps from 2 to 64, both training strategies produced closely matched accuracy as the time steps increased. Accuracy exceeded 99.0% from 4 time steps onward and reached 99.22% at 32 time steps for both methods. At 2 time steps, RE achieved 98.73% accuracy, which outperformed 98.65% obtained with surrogate gradients. Despite the similar precision, training time differed substantially. The training time required by SG method scaled strongly with the number of time steps, increasing from 26.42 seconds at 2 time steps to 136.80 seconds at 64 time steps. In contrast, the training time consumption of our proposed RE method remained nearly constant at about 6.7 seconds across all settings. This yields a 3.9-fold speedup at 2 time steps and roughly a 20-fold speedup at 64 time steps while preserving accuracy. Overall, RE effectively decouples training cost from simulation length, enabling efficient optimization of SBNN over both short and long temporal windows without compromising classification performance.

**Robustness to Noise and Hardware Constraints**

Compared to biological circuits whose computation is inherently robust to noisy, the performance of neural networks would be prohibited by the stochasticity of neuromorphic hardware on which the algorithms are deployed. To bridge this gap and ensure reliable deployment, it is critical to evaluate how spiking models withstand these non-ideal environments. In this section, we assess the resilience of our proposed SBNN under challenging non-ideal conditions. Our evaluation covers two primary dimensions, stochastic perturbations, comprising weight noise, input corruption, as well as threshold variability, and hardware constraints, including performance at small network sizes low-bit precision.

**SBNN Works Robustly in Noisy Environment.** Device inherent noise is a factor that degrades the performance of the deployed algorithm networks. Here we assess the robustness of the proposed SBNN framework to the device noise was assessed by injecting Gaussian perturbations into synaptic weights, input images and neuronal thresholds, as shown in Fig. 2. (B) variant of the SBNN architecture, which was trained with surrogate gradients (SBNN-SG) was used as an ablation in Fig. 2(C) to dissociate the effects of the architecture and the training method.

Under synaptic weight perturbations (Fig. 2 (B)), the SNN baseline showed a pronounced loss of accuracy as the noise increases. Performance dropped from 96.0% under a weight noise with 0.05 Gaussian standard deviation to 29.9% under 0.20 standard deviation Gaussian noise. SBNN-SG markedly improved robustness and yielded accuracy of 98.3% and 63.8% with the corresponding noise levels. Our proposed framework was largely insensitive to weight perturbations and maintained 98.6% accuracy under 0.05 Gaussian standard deviation and 96.5% under 0.20 standard deviation Gaussian noise. At the highest noise level, our SBNN gained about 67% accuracy improvement over SNN baseline and about 33% accuracy growth over SBNN-SG.

For additive input perturbations (Fig. 2 (C)), we compared standard SNN baseline with the proposed framework. At low noise levels, with a Gaussian standard deviation of 0.10 or less, both models achieved similar performance around 97% to 98%. As the input noise increased, the SBNN degraded more slowly than SNN baseline. At a standard deviation of

0.25, the SBNN retained 81.9% accuracy, whereas SNN baseline fell to 69.5%, a difference of roughly 12% accuracy degradation.

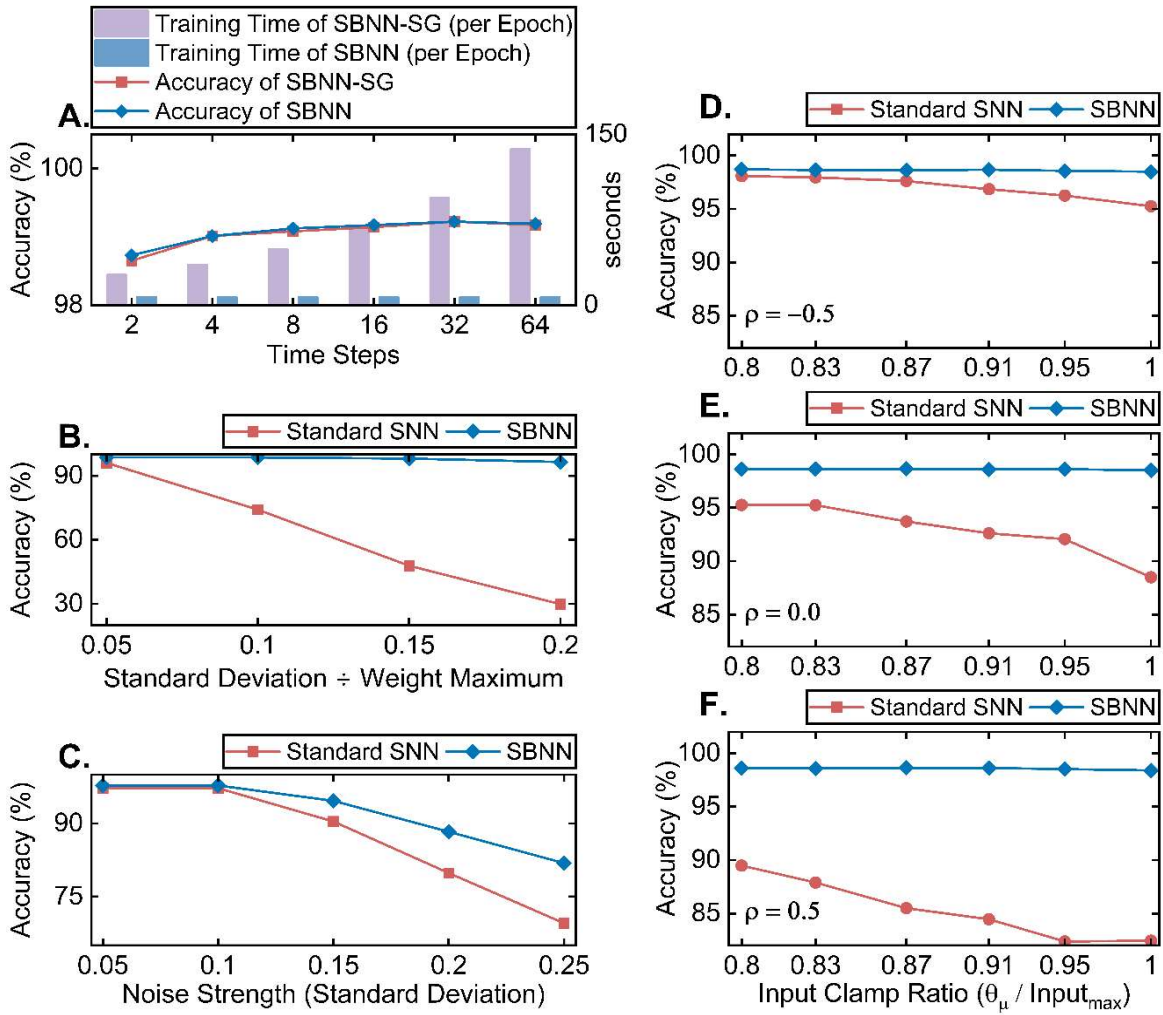

**Fig. 2. Time-Step Scaling and Robustness to Perturbations in SBNN.** Performance of a standard SNN trained with SG and the proposed SBNN trained with the RE method, evaluated on a 784–1000–10 FCN. (A) Accuracy and wall clock training time as a function of time steps for RE and surrogate gradient training. For matched time steps, RE attains comparable accuracy while requiring substantially less training time, and its runtime remains nearly constant as time steps increase. (B) Weight perturbation: During the testing time, additive Gaussian noise is applied to the synaptic weights. The horizontal axis represents the noise strength, which is equal to the standard deviation of the noise divided by the maximum value of the weights. (C) Input perturbation: Gaussian noise is added to input images at test time and the $x$ axis is the noise standard deviation. (D to F) Neuronal-threshold perturbation with input clipping: Gaussian noise is added to neuronal thresholds with scale set by $\log(1 + \exp(\rho))$ while neuron inputs are clipped to a maximum value and the x axis is the ratio of the clipping maximum to the mean threshold. (D) $\rho = -0.5$ (E) $\rho = 0.0$ (F) $\rho = 0.5$

Next, we investigated the combined impact of neuronal threshold perturbations and input range constraints (Fig. 2 (D)), which emulate the device stochasticity and restricted input voltage ranges required in neuromorphic hardware. To faithfully represent the noise of physical devices, which is influenced by the size of devices, different Gaussian noise was applied to the thresholds, parameterized by a mean $\mu$ and a standard deviation $\sigma = \log(1 + e^\rho)$. Simultaneously, to simulate the operating ranges of hardware, the incoming current for each neuron was clipped at a deterministic maximum. This upper bound was characterized by the mean of threshold and a scaling ratio $r_s$, defined as $Input_{max} = \mu/r_s$. The horizontal axis of Fig. 2 (D) depicts the ratio $r_s$. For the SNN baseline, increasing the $r_s$ from 0.8 to 1.0 (thereby tightening the input constraint) led to significant accuracy losses. Specifically, at $\rho = 0$, accuracy deteriorated from 95.3% to 88.5%, and at $\rho = 0.5$, the accuracy fell from 89.5% to 82.4%. Even under low-noise conditions ($\rho = -0.5$), the baseline suffered a performance drop from 98.1% to 95.3%. In contrast, our proposed SBNN framework remained remarkably stable across all tested $\rho$ values and clipping ratios. Accuracies were maintained within a narrow band of 98.4%–98.7% for $\rho \in \{-0.5, 0.0\}$ and 98.4%–98.6% for $\rho = 0.5$, even under the strictest clamping conditions. This stark disparity highlights the advantage of our approach, unlike the baseline, which falters under hardware-imposed constraints, our SBNN exhibits robust adaptability to the combined challenges of limited dynamic ranges and stochastic noise.

Collectively, these results across synaptic, input, and threshold perturbations demonstrate that our framework consistently outperforms the SNN baseline, exhibiting superior robustness to the noise and operating constraints characteristic of neuromorphic hardware.

**Table 2. SBNN Performance in Tiny Architecture.** This table summarizes test accuracy on MNIST for networks with a single hidden layer of width $x$. We report results across hidden-layer sizes from 10 to 100 neurons and compare SBNN with prior compact models under the same 784–$x$–10 topology. $W$ denotes synaptic weight precision in bits. Hidden-layer width modulates model capacity, allowing direct assessment of how accuracy scales with architectural size and numerical precision in small networks.

| Model | W | # of Hidden Neurons | | | | |
|---|---|---|---|---|---|---|
| | | 10 | 20 | 30 | 50 | 100 |
| Westby | 32 | 91.67% | 93.66% | 94.89% | 95.68% | - |
| Dariol | | 85.00% | - | - | - | - |
| SPDNN | | 89.87% | 94.60% | 96.41% | 97.77% | 98.61% |
| SNN | | 89.09% | 93.78% | 95.57% | 97.05% | 98.23% |
| **SBNN (Ours)** | **8** | **90.84%** | **93.49%** | **95.80%** | **96.86%** | **97.87%** |

**SBNN Performed Well in Tiny Architecture.** Here, we evaluated the performance of our SBNN in a parameter-constrained environment. The trained tiny SBNN was evaluated on MNIST dataset. Table 2 reports the test accuracy for compact FCN with a single hidden layer of width $x$, all evaluated under the same 784–$x$–10 topology. We swept the hidden-layer width over $x \in {10,20,30,50,100}$ to examine the accuracy with different hidden layer width $x$ in small networks, where performance is typically sensitive to both model size and weight precision. Importantly, all prior compact models in the table used 32-bit weights, whereas our SBNN was evaluated with only 8-bit weights. In other words, the robustness of our 10 time steps SBNN was tested under a tighter precision constraint. Despite this 4× reduction in weight precision, SBNN achieved consistently high accuracy across all hidden-layer widths. In the most constrained setting ($x = 10$), SBNN reaches 90.84%, essentially matching the 32-bit Westby's model (62) at 91.67% and substantially outperforms Dariol's model (63) at 85.00%. As width increases, the accuracy of SBNN improved smoothly and monotonically, indicating that the model effectively benefits from increased capacity while remaining stable under low-precision quantization. Compared to 32-bit compact baselines, SBNN remains highly competitive even at larger widths. For example, at $x = 50$, SBNN achieves 96.86%, which is close to SPDNN(64) (97.77%), and at $x = 100$ SBNN attains 97.87% compared with SPDNN's 98.61%. While SPDNN yields the highest absolute accuracy at wider settings,

SBNN achieves near the SOTA performance among compact models with significantly reduced numerical precision. This indicates that our model is suitable for edge computing hardware integration.

**Bayesian Uncertainty Estimation with SBNNs**

To evaluate predictive calibration and sampling efficiency of SBNNs, we study uncertainty propagation in SBNNs and its implications for predictive reliability. On one hand, we separated neuronal from synaptic stochasticity and characterized the role of CLT behavior in shaping the distribution of membrane potentials. On the other hand, we measured how sampling efficiency trades off against predictive robustness and compared our SBNN to recent Bayesian baselines.

**Neuronal Uncertainty and Synaptic Uncertainty.** To understand the differences in mechanisms between the threshold stochasticity of our SBNN and the weight stochasticity of standard BNNs, we analyzed the variations in the distributions of membrane potential at a single neuron. We first explored the variations induced by synaptic uncertainty. Under a standard Bayesian formulation, each synaptic weight is modeled as a deterministic mean plus an independent zero-mean random perturbation. Consequently, the membrane potential, which is computed as the weighted sum of inputs, represents a summation of multiple independent random variables. By virtue of CLT, as the number of input connections (fan-in) increases, the distribution of this membrane potential asymptotically converges to a Gaussian. On the other hand, neuronal stochasticity can be viewed as directly imposing uncertainty on the membrane potential threshold, thereby avoiding this degeneration.

In the wide-network and under this scaling, we find that the membrane potential of each neuron admits a simple effective description. The many independent synaptic perturbations entering neuron $j$ aggregate, by the CLT, into an approximately Gaussian membrane potential noise term. The resulting input can be written as

$$x_j^\ell \approx m_j + \eta_j$$

where $m_j$ is the mean membrane potential and $\eta_j$ is an approximately Gaussian random variable with zero mean and a width-independent variance $\sigma_{j,\text{eff}}^2$. Thus, from the neuron's perspective, the detailed microscopic statistics of the incoming synaptic weights are fully summarized by the low-dimensional parameters $(m_j, \sigma_{j,\text{eff}}^2)$.

This representation is naturally equivalent, in distribution, to a model with deterministic synaptic weights and stochastic neuron-wise thresholds. Writing the neuron output as

$$S_j^l = \phi(x_j^\ell)$$

where $\phi$ is a fixed activation function, we can reparameterize the effective noise as a random threshold $\theta_j := -\eta_j$ and express the same stochastic output as

$$S_j^\ell \approx \phi(\bar{x}_j^\ell - \theta_j)$$

where $\bar{x}_j^l := m_j$. In this view, wide networks with synaptic noise are distributionally equivalent, at the output level, to networks with deterministic synapses and neuron-specific Gaussian threshold priors.

Crucially, however, this equivalence holds after the multi-synapse aggregation has already taken place, the CLT compressing synaptic uncertainty into a simple Gaussian perturbation with fixed variance, leaving only a low-dimensional summary of the original weight prior. In contrast, if one directly places a prior on neuron thresholds, its distribution and variance are specified explicitly and do not undergo further averaging or Gaussian as the network width increases. Threshold noise therefore provides an equivalent representation of the synaptic uncertainty while preserving the original prior degrees of freedom and strength, and it is not additionally constrained or regularized by increasing network width.

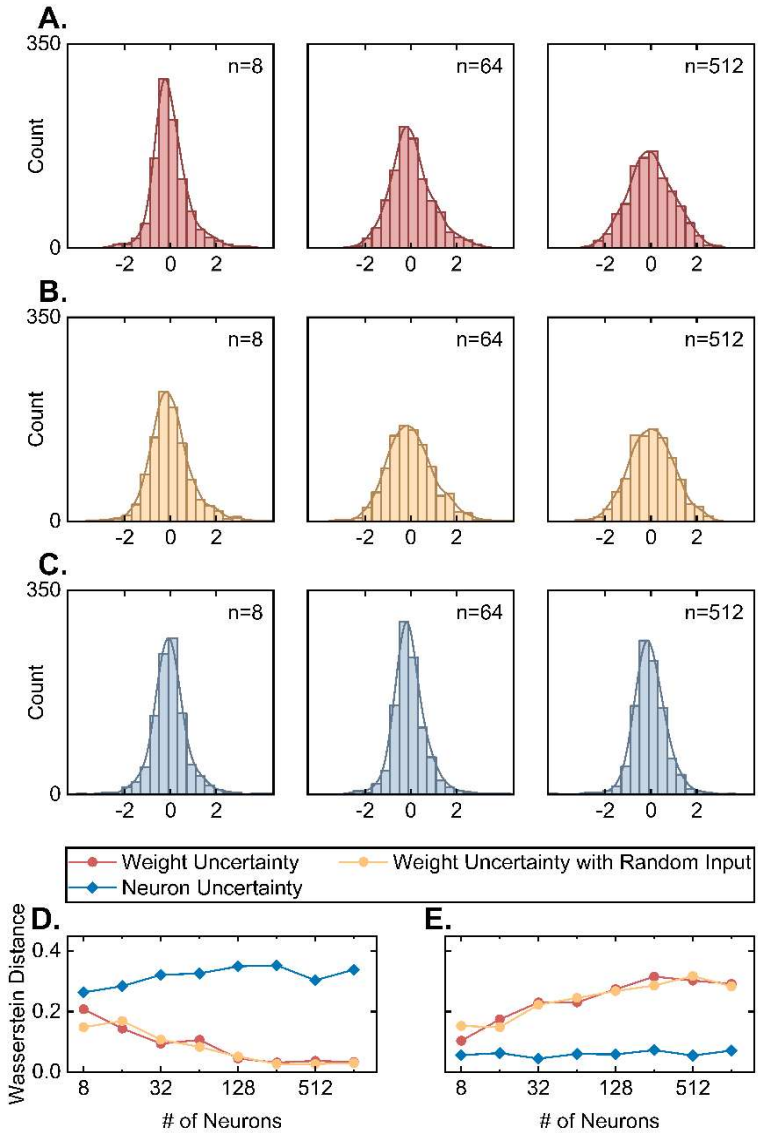

**Fig. 3. Lévy α-stable Distribution of Three Type Uncertainty Injection.** (A) Distribution of membrane potential when the weight following Lévy α-stable distribution (B) Distribution of membrane potential when the weight following Lévy α-stable distribution and the inputs

are fellow Bernoulli distribution. (C) Distribution of membrane potential when the neuronal threshold following Lévy α-stable distribution and the inputs are fellow Bernoulli distribution. (D) Wasserstein-1 distance between the normalized membrane distribution and standard Gaussian distribution. (E) Wasserstein-1 distance between the normalized membrane distribution and standard Lévy α-stable distribution.

Consistent with this prediction, increasing fan-in under synaptic uncertainty produced a systematic decrease in the Wasserstein-1 distance between standardized membrane potential and standard normal, indicating progressive Gaussian with aggregation. In contrast, neuronal uncertainty showed no fan-in dependence in this metric: standardized membrane potentials remained close to the corresponding standardized base distribution across $n$. With fixed activation rate $p$ under He scaling, the membrane potential variance stayed approximately constant, indicating that the separation is driven by distributional shape rather than scale.

Using a truncated, skewed Lévy $\alpha$-stable distribution as a biologically motivated heavy-tailed reference (28), we quantified shape preservation by $d_{\text{ori}} = W_1\big(\text{std}(x), r_{\text{Levy}}\big)$ and Gaussian by $d_N = W_1\big(\text{std}(x), \mathcal{N}(0,1)\big)$. As is shown in Fig. 3 (A), under synapse weight sampling, Lévy-like structure degraded with increasing fan-in. In this situation, $d_{\text{ori}}$ rose from 0.103 at $n = 8$ to 0.231 at $n = 64$ and 0.303 at $n = 512$. The same degradation persisted in Fig. 3 (B), when inputs were randomized, indicating that input stochasticity does not rescue the target heavy-tail statistics under weight uncertainty. By contrast, neuron noise preserved the Lévy-like shape across the same range, with consistently low $d_{\text{ori}}$ (0.056, 0.060, 0.055 for $n = 8, 64, 512$). Consistently, $d_N$ decreased with fan-in for weight uncertainty but remained comparatively elevated for neuronal uncertainty, reflecting sustained non-Gaussian. The same results are also observed in additional noise families, including exponential family and non-exponential family distribution. Results for additional noise families (Gaussian, exponential, gamma, symmetric Pareto, and Student-t) are provided in the Supplementary Materials (Figs. S2–S8).

**Well Calibration Performance of SBNN in Different Architecture.** Here, we assess the calibration performance of our SBNN on MNIST classification tasks. All models were trained with 16 time steps. Because the proposed SBNN is able to produce one independent Monte Carlo sample per time step, in the inference stage, the number of time steps and the number of Monte Carlo (MC) runs can be balanced under a fixed sampling budget.

In Fig. 4 (A) we fixed the total sampling budget (time steps multiplies MC runs) to 320. Varying the number of time steps revealed a best performance trade-off operating regime in which both predictive uncertainty and likelihood-based calibration improved. As the number of time steps was reduced from 320 to 8, the negative log likelihood (NLL) decreased monotonically and reached its minimum of 0.03179 at 8 time steps, while the mean predictive entropy remained low at 0.07509. A similarly stable behavior was observed at 16 time steps, with an NLL of 0.04223 and entropy of 0.01497. In contrast, further reducing the time steps led to a sharp degradation in both NLL and entropy, while increasing the time steps caused a degrading NLL alone. These results indicate that, under a fixed sampling budget, excessively few time steps or few MC runs fail to capture the posterior induced predictive distribution, whereas an intermediate range yields lower NLL together with reduced classification entropy, consistent with improved calibration.

Fig. 4 (B) evaluated classification performance and calibration on a 784-400-400-10 architecture for models using 8 or 16 time steps in the best performance trade-off operating regime. Across different MC runs, 8 time steps achieved essentially the same accuracy as 16 time steps while consistently improving NLL. With 16 Monte Carlo runs, the 8-time step model reached an accuracy of 0.9913 and an NLL of 0.03274, while the 16-time step model achieved an accuracy of 0.9912 and an NLL of 0.04245. The same trend held at low sampling budgets. With a single Monte Carlo run, the 8-time step model achieved an NLL of 0.04413 whereas the 16-time step model yielded 0.0653. This comparison shows that reducing inference to 8 time steps preserves classification accuracy and improves likelihood-based calibration.

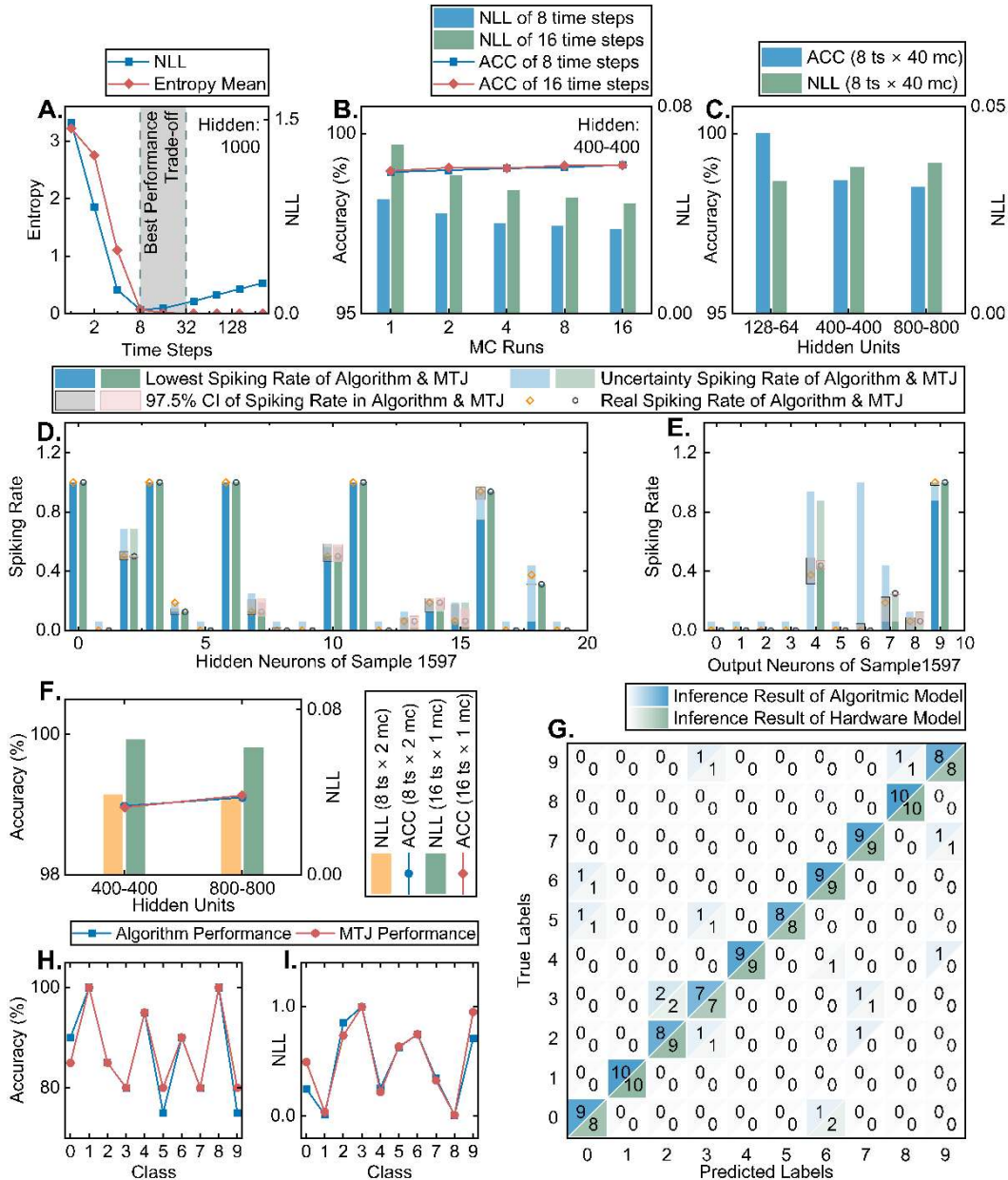

**Fig. 4. Calibration and Device Validation of Our SBNN on MNIST.** (A) Under a fixed sampling budget of 320 total steps, varying the number of time steps reveals an optimal regime in which both negative log likelihood (NLL) and mean predictive entropy are minimized, indicating improved calibration. (B) On a 784–400–400–10 architecture, inference with 8 time steps matches the accuracy of 16 time steps while yielding lower NLL across various Monte Carlo runs. (C) At 8 time steps, increasing hidden-layer width improves accuracy and reduces NLL, and comparable calibration is obtained with as few as two Monte

Carlo runs relative to 40-run averaging, consistent with efficient uncertainty estimation from independent samples per step. (D) Hidden layer spiking rates for MNIST sample 1597 predicted by the algorithm and measured on MTJ hardware. Device rates were obtained by sweeping the input voltage. For each voltage, we pulsed the MTJ, inferred the post pulse state from the resistance, and counted switching events as spikes. The minimum spiking rate was estimated from voltages with switching probability above 99%. Uncertainty was quantified from all voltages with switching probability above 1%. Error bars denote 97.5% confidence intervals. (E) Output layer spiking rates for MNIST sample 1597 predicted by the algorithm and measured on MTJ hardware. Hidden layer spike activity was projected to the output layer using the learned weights to obtain each neuron input. After amplitude scaling, the corresponding value was applied as the MTJ input voltage and the device state was read out to estimate spiking rates. (F) Accuracy and NLL of same step budget (total 16 steps, equal to times steps multiple mc runs) across two architecture. (G) Confusion matrix comparing inference by the algorithm model and the MTJ hardware. Trained parameters were deployed on hardware and inputs were scaled using the learned neuron threshold. The same test set was evaluated by the algorithm and by the device. Results are shown as a split cell matrix, with the upper left triangle reporting algorithm predictions and the lower right triangle reporting hardware predictions. (H) Per class accuracy using eight time steps and two Monte Carlo samples, comparing algorithm accuracy and device accuracy. (I) Per class NLL under the same eight step, two sample setting, quantifying calibration differences between algorithm and MTJ hardware.

In Fig. 4 (C) we studied the role of model size at 8 time steps and reported results averaging over 40 MC runs. Models with smaller hidden layers underperformed, as the 128-64 network achieved an accuracy of 0.9868 with an NLL of 0.04354. Increasing capacity to 400-400 improved accuracy to 0.9907 and reduced NLL to 0.0322, and the 800-800 model further improved to 0.9917 accuracy with an NLL of 0.0307. Notably, the 400-400 model retained comparable calibration even when the number of Monte Carlo runs was reduced from 40 to 2, achieving 0.9901 accuracy with an NLL of 0.038. Given that each time step corresponds to an independent sample, this suggests that a compute budget comparable to roughly 16 samples can already yield high quality uncertainty estimates, enabling efficient inference without sacrificing probabilistic calibration.

We next explored the advantages of our SBNN over representative BNN baselines on three multilayer perceptron architectures (128–64, 400–400, 800–800), as summarized in Table 2. In these comparisons, we adopt 8 time steps and 40 MC runs because of the previously found best performance trade-off operating regime.

To this end, Table 3 compares SBNN with the recent SUQ model (65), implicit generative-prior (66) BNNs, and collapsed-variational MF-VI variants (67) on MNIST across three MLP architectures (128–64, 400–400, 800–800), reporting test accuracy and negative log-likelihood (NLL) as a standard proxy for predictive calibration.

Compared to other BNN models, SBNN consistently improves both predictive accuracy and probabilistic calibration (as measured by negative log-likelihood, NLL) across three network architectures. For the compact 128–64 model, SBNN reaches 98.68% accuracy with NLL = 0.044, exceeding the SUQ model (65) (LA/MFVI; 98.00%, NLL 0.063/0.062) and yielding a 0.019 NLL reduction relative to the best streamlined baseline. Scaling to 400–400 hidden units, SBNN achieves 99.07% accuracy and NLL = 0.032, outperforming implicit-prior BNN baselines (VBNN/SVBNN/NA-EB (66)) and collapsed-variational MF-VI variants (67), including the strongest competitor trained with a large batch (NLL = 0.041), corresponding to a further 0.09 decrease in NLL while also improving accuracy (best competing accuracy 98.76%). The same trend holds for the largest 800–800 model, where SBNN attains 99.17% accuracy and the lowest NLL = 0.031, surpassing the best MF-VI comparator (98.80%, 0.042). Collectively, these results indicate that SBNN delivers better-calibrated predictive distributions, robustly across model capacity.

Furthermore, we evaluated the robustness of the proposed SBNN to out-of-distribution (OOD) data. We evaluated the performance of SBNN and SNN trained on the MNIST dataset on the rotated version of the test set and compared them with a trained SUQ model (65). Compared to other models, our SBNN consistently maintains low NLL across MNIST images with multiple rotation angles, demonstrating that SBNN reduces overconfidence in OOD data, as shown in supplementary materials Fig. S1.

**Deployment of SBNN Parameters on Device**

We evaluated the fidelity of SBNN deployment on MTJ hardware by transferring the trained parameters to devices and testing on a randomly selected balanced subset of the MNIST test set (100 images, 10 images per class). Based on measured MTJ switching characteristics, we calibrated an input modulation scheme that maps neuronal inputs to device drive voltages for device-level inference.

**Table 2. Calibration Performance of SBNN.** Comparison of test accuracy (↑) and negative log-likelihood, NLL (↓), which is a standard proxy for predictive calibration, for SBNN and BNN baselines across three multilayer perceptron architectures (hidden units including three scales, 128–64, 400–400, 800–800). Baselines include SUQ model (65) (LA and MFVI), implicit generative-prior (66) BNNs (VBNN, SVBNN, NA-EB), and collapsed-variational (67) MF-VI variants (CM-, CV-, CMV-MF-VI), as well as CM-MF-VI (4000batch) denotes the same approach trained with batch size 4000. SBNN achieves the lowest NLL and the highest accuracy in every architecture, demonstrating consistently improved calibration under scaling.

| Hidden Units | Model | Accuracy (%) ↑ | NLL ↓ |
|---|---|---|---|
| 128-64 | LA SUQ † | 98.00 | 0.063 |
|  | MFVI SUQ † | 98.00 | 0.062 |
|  | **Ours** | **98.68** | **0.044** |
| 400-400 | VBNN | 98.64 | 0.118 |
|  | SVBNN | 98.60 | 0.144 |
|  | NA-EB | 98.76 | 0.057 |
|  | CM-MF-VI | 98.66 | 0.047 |
|  | CV-MF-VI | 97.87 | 0.068 |
|  | CMV-MF-VI | 98.55 | 0.049 |
|  | CM-MF-VI (4000batch) | 98.69 | 0.041 |
|  | **Ours** | **99.07** | **0.032** |
| 800-800 | CM-MF-VI | 98.58 | 0.048 |
|  | CV-MF-VI | 97.91 | 0.066 |
|  | CMV-MF-VI | 98.40 | 0.052 |
|  | CM-MF-VI (4000batch) | 98.80 | 0.042 |
|  | **Ours** | **99.17** | **0.031** |

† indicates that the model has been calibrated for temperature scaling parameters on the validation set.

To ensure compatibility between the algorithmic SBNN and the MTJ hardware we rescaled the inputs of each neuron during inference. For a neuron with learned threshold $\mu$ and input $x_{in}$, the modulated input was given by

$$x_{mod} = \frac{0.4997 \cdot x_{in}}{\mu} \qquad (9)$$

This mapping aligned the effective neuronal threshold with the measured switching characteristic of the MTJ devices.

We trained an SBNN (784-20-10, MLP) on the full MNIST training set and deployed the learned parameters to the MTJ hardware for device level inference. For evaluation we randomly drew 10 images per class from the MNIST test set which gave 100 samples in total, as is shown in the supplementary materials Fig. S12. Each image was converted into a 16 step Poisson spike train. After training we extracted the synaptic weights and neuronal thresholds. The modulated inputs were converted to MTJ drive voltages according to the Eq. (9). Hidden layer activity was emulated by applying these voltages to MTJs and reading out the post resistance to determine the device state at each step. The resulting hidden layer spike trains were linearly combined with the learned weights to obtain the inputs to the output layer. Output neurons were implemented in the same way on MTJs and their spiking rates over the 16 steps formed the final network readout.

Fig. 4 (D) and Fig. 4 (E) analyze inference for an example sample ( the $1597^{th}$ sample in MNIST test set which belongs to class 9). We plot the spike rate of each neuron in the hidden and output layers for both the algorithmic model and the MTJ based implementation. Individual spike times are not perfectly matched between the two realizations. In contrast the mean spike rates of corresponding neurons differ only slightly, and the minimum attainable rate of each neuron is identical in the two cases. The main visible difference is the width of the uncertainty bands for some neurons. We attribute this effect to imperfect calibration of the neuron specific standard deviation in the hardware model. These results indicate that the MTJ hardware preserves the deterministic component of the rate code and that intrinsic thermal noise drives fluctuations around these rates that are consistent with approximate Bayesian inference. See the supplementary materials Fig S9-S10 for the details of other example samples.

In previous sections, we found that although the models are trained in 16 time steps, their NLL performs better in 8 time steps. This discovery prompted us to investigate the splitting the inference of 16 independent spikes into two distinct 8-time-step MC samples, as is shown in Fig. 4 (F). Across two different architectures, the accuracy of the 16-time-step model was very similar to that of the 8-time-step model, but the calibration performance (NLL) of the 8-time-step model was significantly better. We attributed this improvement in calibration performance to the benefits of twice mc samples. As a result, in the hardware implementation, we used 16 spikes for inference and separated them into two samples for calibration analysis.

Fig. 4 (G) compares the confusion matrices obtained from the algorithmic SBNN and from the MTJ hardware. Entries from the algorithm appear in the upper left half of each cell

and entries from the hardware appear in the lower right half. These numbers represented the count of each possible prediction case (the predicted label in horizontal axis versus true label in vertical axis). The overall classification accuracy is identical for the two implementations. Most errors occur between visually similar classes which suggest that deployment on hardware introduces minimal additional degradation.

Fig. 4 (H) reports the class-wise classification accuracy for the algorithmic SBNN and the MTJ-based deployment, evaluated on 100 randomly selected MNIST test samples. The MTJ implementation closely tracks the algorithm across all classes. Small discrepancies are observed mainly for classes 0, 5, and 9, while the remaining classes show identical or nearly identical accuracy, indicating that mapping trained parameters to the device introduces minimal class-dependent degradation.

Fig. 4 (I) compares the class-wise NLL between the algorithmic model and the MTJ hardware, providing a calibration-sensitive metric that reflects both prediction correctness and confidence. The overall NLL profile is largely preserved across classes, demonstrating similar uncertainty calibration after deployment. Modest NLL increases are visible for some classes (notably 0 and 9), consistent with the wider uncertainty bands observed for a subset of neurons in Fig. 5(D, E). For all the incorrect samples, the predicted results of MTJ were consistent with the results predicted by the algorithmic model, proving that the errors stemmed from the recognition capabilities of models, not from implementation errors (as is shown in the supplementary materials Fig. S11).

Taken together, Fig. 4 (D–I) demonstrate that the MTJ-based implementation not only reproduces the deterministic aspects of the algorithmic model (confusion matrix and per-class accuracy) but also replicates the probabilistic calibration behavior captured by the algorithmic model (NLL). Although individual spike times are not expected to match exactly, the close agreement in class-wise accuracy and NLL indicates that intrinsic device stochasticity drives fluctuations around nearly identical mean firing rates in a manner consistent with approximate Bayesian inference. Therefore, our calibration-and-deployment approach enables leveraging physical randomness in MTJ devices to realize intrinsic Bayesian inference with negligible additional accuracy loss, and the remaining errors are primarily limited by the current network capacity and are expected to decrease with moderate scaling of network width or depth.

**Discussion**
In this work, we reframe the stochastic switching in MTJs as a computational resource for neuromorphic computing, rather than an implementation error. We presented a co-design architecture across hardware and software where intrinsic device randomness was modeled as neuronal uncertainty. In this framework, physically stochastic neurons are combined with variational inference during training so that noise distributions are learned as part of the model. During inference, probabilistic computation emerges through sampling at the neuronal threshold. In this way, the intrinsic device noise serves as the random source required by Bayesian inference.

This design was inspired by evidence that biological circuits exhibit trail-to-trail variability, which relies on the neuronal intrinsic noise (26)(27). Current neuroscience theories suggest that such variability supports probabilistic computation, even Bayesian inference, by enabling sampling-based representations of latent uncertainty (31). By assigning uncertainty to the neuron, our framework is able to represent various distributions that have been observed in the biological brains. This emulation of stochastic firing patterns mirrors the behavior of biological neurons more closely than deterministic models or standard BNNs.

Implementing Bayesian inference/BNNs on conventional deterministic hardware often incurs substantial overhead due to sampling (68), while the neuromorphic hardware is implemented on the low precision systems to avoid errors (39), such as weight value drift (69). Beyond BNNs, the proposed SBNN achieves competitive accuracy to the full-precision SOTA models even applied on the low precision systems. In addition, SBNN offers well performance in which the number of parameters and model size are limited. This capability reduces the memory and storage requirements typically associated with probabilistic models and facilitates deployment in resource-constrained hardware environments.

Our framework not only performs well on accuracy but also endows the network with high robustness against external and internal perturbations, such as input signal and resistance variabilities. The training process takes the inherent device noise into consideration and makes the network remain stable under input noise and reduced supply voltage. This robustness extends to strong perturbations in weight values and neural activities, indicating that embracing noise during optimization can mitigate the effects of physical nonidealities that are otherwise difficult to control in emerging hardware.

A critical outcome of this framework is the achievement of full-precision sampling capabilities using low-precision devices. Previous BNNs typically embed stochasticity within the synapses, but this creates a fundamental mismatch when applied to low-precision hardware. For example, in binary regimes ($w \in \{-1, +1\}$), noise-induced fluctuations result in abrupt, large-scale state transitions (e.g., flipping from $-1$ to $+1$) rather than subtle probabilistic adjustments. Our SBNN resolves this by shifting the stochastic source to the neuronal threshold, effectively decoupling sampling precision from weight quantization. By injecting continuous noise from the physical world into the neuron, we bypass the limitations of quantized weights and achieve high-fidelity, full-precision sampling. Consequently, this architecture significantly minimizes hardware complexity and maximizes compactness, making it exceptionally well-suited for the stringent resource constraints of emerging edge computing systems.

On the other hand, hardware experiments confirm the physical realizability of this framework without performance degradation. We mapped trained thresholds to the device by scaling the input voltage values and performed inference. The computations conducted on the fabricated devices match the accuracy observed in algorithmic inference and show no visible additional loss after mapping. This agreement indicates that MTJ switching acts effectively as a stochastic neuron and that Bayesian inference is achievable without requiring safety calibration or extensive error-correction circuits.

The broader implication is that probabilistic neuromorphic systems can be built by embracing intrinsic device noise at the neuronal level. This architecture provides a novel method for injecting randomness that complements existing synaptic approaches. While current experiments focus on small to medium networks, the inherent robustness of the model offers a promising pathway for scaling to larger noise-sensitive networks. A key advantage of our framework is its ability to implant distinct noise distributions that can be obtained from the intrinsic physical characteristics of various devices. This flexibility implies that the approach is not limited to MTJs but can be generalized to other emerging technologies. Future work will explore combined sources of uncertainty in both neurons and synapses and extend this framework to other emerging memory technologies that exhibit reproducible switching statistics.

**Materials and Methods**

**Device fabrication**

The MTJ devices studied here were patterned from a (thermal silicon oxide substrate) /Ta(5)/CuN(20)/Ta(5)/[Pt(2.5)/Co(1)/Ta(0.5)]$_9$/Pt(2.5)/Co(1)/Ta(1)/ $Co_{40}Fe_{40}B_{20}$(0.9)/MgO(0.85)/$Co_{20}Fe_{60}B_{20}$(1.1)/Ta(0.5)/Co(0.3)/ [Pt(1.5)/Co(0.4)]$_2$/Ru(0.85)/[Co(0.5)/Pt(1.5)]$_3$/Ru(5) multilayer, deposited by using a Singulus ROTARIS physical vapor deposition system (thickness in nm). The 0.5 nm thick Ta dusting layer was inserted between $Co_{20}Fe_{60}B_{20}$ and Co(0.3)/[Pt(1.5)/Co(0.4)]$_2$, layers to enhance the perpendicular magnetic anisotropy (PMA) of the top reference layers. Here, a CuN layer wasused as a bufer layer for the growth of the MTJ stack. The patterned electrodes, Ti(10nm) and Au(100nm) were fabricated using a lift-off process.

**MTJ-based Stochastic Neuron Model and Bayesian Threshold Dynamics**

The low-resistance parallel state and high-resistance anti-parallel state are stablized by an energy barrier $(\Delta E)$ due to magnetic anisotropy in an MTJ, as is shown in the Fig. 1(B). Due to thermal noise, the switching between parallel and anti-parallel state under applied signals shows stochasticity. Experimental and simulation results show that the switching probability increases with the input current or voltage pulse amplitude with a sigmoid relation(70), as shown in Table 1 (C) . This indicates a random threshold value $V_C$. In every attempt to switch the device state by applied pulse signal, the threshold $V_C$ samples from its underlying distribution. If the applied pulse amplitude surpasses the sampled $V_C$ value, the device state gets switched, otherwise the device remains its original state. After each switching attempt, the device state is reset to the original state. In order to produce the sigmoid relation between switching probability and input pulse amplitude that is observed in this process, a Gaussian distributed threshold $V_C$ is required, since the differentiation of sigmoid relation brings Gaussian relation, as is shown in the Equation (2).

**Network Architecture and Noise Models**

We consider a feedforward layer with input activations $S^{\ell-1} \in \mathbb{R}^n$ and pre-activations

$$x_j^l = \sum_{i=1}^{n} w_{ij}^\ell S_i^{\ell-1} \qquad (10)$$

Synaptic weights follow

$$w_{ij}^l = \mu_{ij}^l + \epsilon_{ij}^l, E[\epsilon_{ij}^l] = 0, Var(\epsilon_{ij}^l) = (\sigma_{ij}^l)^2 \qquad (11)$$

where the $\epsilon_{ij}^l$ are mutually independent and independent of $S^{l-1}$. Substitution into Eq. (10) yields

$$x_j^l = \sum_{i=1}^{n} \mu_{ij}^l S_i^{l-1} + \sum_{i=1}^{n} \epsilon_{ij}^l S_i^{l-1} \qquad (12)$$

We define the mean and noise contributions

$$M_j(x^l) := \sum_{i=1}^{n} \mu_{ij}^l S_i^{l-1}, \quad N_j(x^l) := \sum_{i=1}^{n} \epsilon_{ij}^l S_i^{l-1} \tag{13}$$

By construction,

$$\mathbb{E}\big[N_j(x^l) \mid S^{l-1}\big] = 0, \quad \mathbb{E}\big[x_j^l \mid S^{l-1}\big] = M_j(x^l) \tag{14}$$

and averaging over $S^{l-1}$ defines the neuron-specific mean $m_j := \mathbb{E}[M_j(x^l)]$.

Conditional on $S^{l-1}$, the terms

$$X_i := \epsilon_{ij}^l S_i^{l-1} \tag{15}$$

are independent, zero-mean random variables with conditional variances

$$\mathrm{Var}(X_i \mid S^{l-1}) = (S_i^{l-1})^2 (\sigma_{ij}^l)^2 \tag{16}$$

Assuming finite second moments and that no single term dominates the total variance, the Lindeberg–Feller central limit theorem implies that

$$\frac{N_j(x^l)}{\sqrt{\sum_{i=1}^{n} (S_i^{l-1})^2 (\sigma_{ij}^l)^2}} \xrightarrow{d} \mathcal{N}(0,1) \text{ as } n \to \infty \tag{17}$$

Equivalently, for large $n$,

$$N_j(x^l) \mid S^{l-1} \approx \mathcal{N}\Big(0, \sum_{i=1}^{n} (S_i^{l-1})^2 (\sigma_{ij}^l)^2\Big) \tag{18}$$

To obtain a width-independent effective variance, we adopt the usual wide-network scaling

$$(\sigma_{ij}^l)^2 = \frac{(\tilde{\sigma}_{ij}^l)^2}{n} \tag{19}$$

with $\tilde{\sigma}_{ij}^l = O(1)$ and $\mathbb{E}[(S_i^{l-1})^2]$ bounded. Under these conditions, the sum in Eq. (18) concentrates, by the law of large numbers, around a finite constant

$$\sigma_{j,\mathrm{eff}}^2 := \lim_{n \to \infty} \sum_{i=1}^{n} (S_i^{l-1})^2 (\sigma_{ij}^l)^2 \tag{20}$$

and its relative fluctuations vanish as $n \to \infty$. Thus, in the wide-network limit,

$$x_j^l \approx m_j + \eta_j, \quad \eta_j \sim \mathcal{N}(0, \sigma_{j,\mathrm{eff}}^2) \tag{21}$$

providing a neuron-wise Gaussian approximation to the aggregated synaptic noise. We let the neuron output be

$$S_j^l = \phi(x_j^l) \tag{22}$$

for a fixed activation function $\phi$. Using Eq. (21) and defining $\bar{x}_j^l := m_j$ and $\theta_j := -\eta_j$, we obtain the equivalent representation

$$S_j^l \approx \phi(\bar{x}_j^l + \eta_j) = \phi(\bar{x}_j^l - \theta_j) \tag{23}$$

Thus, in the wide-network regime under the scaling of Eq. (18), a model with microscopic synaptic weight noise is distributionally equivalent, at the level of neuron outputs, to a model with deterministic weights and a Gaussian prior on neuron-specific thresholds $\theta_j$. Because $\theta_j$ is a single random variable per neuron and is not scaled with $n$, its variance does not diminish with increasing network width, and the resulting output stochasticity remains finite in the limit.

**Limited-Precision Synaptic Weights**

In the spiking-Bayesian (SB) neuron layer, we employ low-precision quantized synaptic weights, using either 1-bit (binary) or 8-bit representations depending on the experiment. This choice yields several hardware-level benefits. First, 1-bit and 8-bit weights reduce memory footprint and data-movement energy by up to 32× and 4×, respectively, relative to 32-bit floating-point parameters. This compression allows complete weight maps to reside in bit-limited on-chip SRAM, resistive RAM (RRAM), or magnetic-tunnel-junction (MTJ) arrays, thereby increasing effective model density.

Second, in the binary configuration, the $\{\pm 1\}$ format matches the native behaviour of MTJ devices, which switch between two stable resistance states. Deterministically driven MTJs then act as binary synapses, whereas stochastically driven devices implement SB neurons. In the multi-bit configuration, 8-bit quantization still preserves low-precision storage while relaxing the constraints on representational capacity, and can be realised via multi-level cells or bit-sliced arrays.

Third, local learning rules intrinsic to SBNNs help compensate for the loss of numerical precision, enabling accuracies that approach those of higher-precision networks while maintaining the energy and area advantages of low-bit storage. Encoding parameters directly in 1-bit or 8-bit form also avoids the accuracy–energy trade-offs associated with fine-grained multi-bit MRAM calibration and provides cell-level reconfigurability, which a single MTJ array can be reassigned between synaptic and neuronal roles during inference, improving utilization of scarce on-chip resources.

**Rate-Estimation Training Algorithm**

Conventional training strategies for spiking neural networks (SNNs), including surrogate-gradient backpropagation and ANN-to-SNN conversion, have well-known limitations. Surrogate-gradient methods depend on hand-crafted gradient approximations, which can distort the underlying optimization objective. Conversion-based methods typically require hundreds of simulation time steps and weight normalization to maintain accuracy. We propose Rate-Estimation (RE) training, which preserves the direct end-to-end optimization of surrogate-gradient approaches while achieving the low latency and hardware efficiency characteristic of ANN-to-SNN conversion. During training, we operate purely in the rate domain, where the firing probabilities $P$ at each layer are given by the outputs $\hat{p}$ of the previous layer, and no spike sampling is performed.

For neuron $i$ in layer $\ell$, the empirical output spiking rate over $T$ discrete time steps is:

$$P_{out} = \frac{\sum_{t=1}^{T} \vartheta(H_t, \theta_t)}{T} \qquad (24)$$

where $X_{in}(t)$ is the total synaptic drive, $\theta$ is the (possibly time-varying) threshold, and $\vartheta(\cdot)$ is the Heaviside step function. Because the step is non-differentiable, direct numerical optimisation of Eq. (24) is intractable.

In Spiking Bayesian Neural Networks (SBNNs), synaptic weights $w_k$ are 8-bit valued and constrained to the interval $[-1,1]$. At time $t$, the total synaptic input to a neuron can be decomposed into excitatory and inhibitory contributions. Let $X$ denote the summed excitatory input and $Y$ the summed inhibitory input, the net input being then,

$$D = X - Y \tag{25}$$

The probability that the net input is positive is:

$$P_{\text{pos}} = \Pr(D > 0) \tag{26}$$

Let $p_k$ be the Bernoulli probability that synapse $k$ emits a spike in the current window. For synapses with positive weights ($w_k > 0$), the contribution to the excitatory input is $w_k$ whenever a spike occurs. For synapses with negative weights ($w_k < 0$), the magnitude $|w_k|$ contributes to the inhibitory input. Under this parameterization, the mean and variance of $D$ are:

$$\begin{aligned}
\mu_D &= \mu_X - \mu_Y, \quad \sigma_D^2 = \sigma_X^2 + \sigma_Y^2 \\
\mu_X &= \sum_{w_k > 0} w_k p_k, \quad \sigma_X^2 = \sum_{w_k > 0} w_k^2 p_k (1 - p_k) \\
\mu_Y &= \sum_{w_k < 0} |w_k| p_k, \quad \sigma_Y^2 = \sum_{w_k < 0} w_k^2 p_k (1 - p_k)
\end{aligned} \tag{27}$$

Assuming that presynaptic spikes are independent across synapses, the Lyapunov condition is satisfied and the net input $D$ is well approximated by a normal random variable with mean $\mu_D$ and variance $\sigma_D^2$. The Gaussian cumulative distribution function appearing in the definition of $P_{\text{pos}}$ has no elementary closed form, so we approximate it with a scaled logistic function $\Phi(z) \approx \sigma(kz)$, where $\sigma(\cdot)$ is the logistic sigmoid and $k \approx 1.716$. This yields

$$P_{\text{pos}} \approx \sigma\left(1.716 \cdot \frac{\mu_D}{\sigma_D}\right) \tag{28}$$

and, more generally, for a neuron with mean threshold $\theta$,

$$P_{\text{out}} = \Pr(D > \theta) \approx \sigma\left(1.716 \cdot \frac{\mu_D - \theta}{\sigma_D}\right) \tag{29}$$

These expressions provide closed-form, differentiable rate estimates for every neuron in $O(m)$ time for an $m \times n$ layer, without requiring a temporal simulation loop during training. As a result, SNNs with low-precision synapses in $[-1,1]$ can be optimized end-to-end using standard stochastic gradient descent by replacing the discontinuous spike nonlinearity with the logistic surrogate above. After training, the learned parameters can be transferred directly to a rate-coded SNN for inference, without any additional rescaling or calibration.

**Rate-Based Convolutional Estimator for Spiking Neurons.** We model the pre-threshold membrane current at each output location as a weighted sum of independent Bernoulli inputs. Let $P \in \mathbb{R}^{B \times C_{\text{in}} \times H \times W}$ denote the input firing probabilities, $W \in \mathbb{R}^{C_{\text{out}} \times C_{\text{in}} \times k_H \times k_W}$ the convolution kernels, and $\theta \in \mathbb{R}^{C_{\text{out}} \times H_{\text{out}} \times W_{\text{out}}}$ a (possibly) spatially varying threshold. Under the independence assumption, the total current $D$ over a receptive field has mean and variance

$$\mu_D = P * W, \qquad \sigma_D^2 = \bigl(P \odot (1 - P)\bigr) * (W \odot W) \tag{30}$$

where $*$ denotes convolution and $\odot$ denotes elementwise (Hadamard) multiplication. For sufficiently many inputs, the central limit theorem implies that the

standardized current is approximately Gaussian, and the firing probability can be written as

$$\Pr(D > \theta) = \Phi\left(\frac{\mu_D - \theta}{\sigma_D}\right) \tag{31}$$

where $\Phi$ is the Gaussian cumulative distribution function.

In practice, we approximate this probit nonlinearity with a scaled logistic (probit–logit approximation) and use

$$\hat{p} = \sigma\left(k \cdot \frac{\mu_D - \theta}{\sigma_D + \varepsilon}\right) \tag{32}$$

where $\sigma(\cdot)$ is the logistic sigmoid, $k \approx 1.716$ is a fixed scaling factor, and $\varepsilon > 0$ is a small constant for numerical stability. This yields a differentiable estimator of output firing rates that can be combined with standard loss functions for end-to-end training. During backpropagation, gradients with respect to $\mu_D$ are computed using standard (or transposed) convolutions. For numerical stability, we treat $\sigma_D$ as a non-trained value, which leads to the following approximate derivatives:

$$\frac{\partial \hat{p}}{\partial \mu_D} \approx \frac{k}{\sigma_D + \varepsilon} \hat{p}(1-\hat{p}), \qquad \frac{\partial \hat{p}}{\partial \theta} \approx -\frac{k}{\sigma_D + \varepsilon} \hat{p}(1-\hat{p}) \tag{33}$$

At test time, the learned parameters $(W, \theta)$ are transferred to a multi-timestep SNN with the same convolutional structure. As the number of simulation steps $T$ and the number of input spikes $S$ increase, the empirical firing rate of the spiking network converges to $\hat{p}$, ensuring consistency between the trained rate model and the deployed spiking dynamics.

**Bayesian Interpretation of the SBNN Forward**

For a spiking Bayesian (SB) neuron, we consider the instantaneous input current at a single time step

$$D = \sum_{i=1}^{n} w_i X_i, \qquad X_i \sim \text{Bernoulli}(x_i) \tag{34}$$

where $w_i \in [-1, 1]$ are quantized synaptic weights and $x_i \in [0, 1]$ denotes the presynaptic firing probability. In our implementation, $w_i$ is represented with $b$-bit precision (we use 1-bit, i.e. binary $w_i \in \pm 1$, and 8-bit settings in the experiments, respectively).

Conditioned on the input probability vector $\mathbf{x} = (x_1, \ldots, x_n)$, the mean and variance of $D$ are

$$\mu_D = \mathbb{E}[D \mid \mathbf{x}] = \sum_i w_i x_i, \qquad \sigma_D^2 = \text{Var}(D \mid \mathbf{x}) = \sum_i w_i^2 x_i (1 - x_i) \tag{35}$$

A spike is emitted when the current exceeds a stochastic firing threshold $\theta$,

$$S = \mathbf{1}[D \geq \theta] \tag{36}$$

We denote by $q_\phi(\theta)$ a variational distribution over thresholds, with cumulative distribution function $F_{q_\phi}$. Approximating the input by its conditional mean $D(\mathbf{x}) = \mu_D$, the single step firing probability is:

$$\begin{aligned}
p_{\text{out}}(\mathbf{x}) &= \Pr(S = 1 \mid \mathbf{x}) \\
&= \mathbb{E}_{\theta \sim q_\phi}[\mathbf{1}[D(\mathbf{x}) \geq \theta]] \\
&= \int_{-\infty}^{D(\mathbf{x})} q_\phi(\theta)\, d\theta \\
&= F_{q_\phi}(D(\mathbf{x}))
\end{aligned} \quad (37)$$

This expression shows that the neuron's firing rate is equal to the probability mass assigned by $q_\phi$ to thresholds less than or equal to the effective input current $D(\mathbf{x})$. In other words, the neuron implements a probabilistic threshold test whose uncertainty is governed by $q_\phi(\theta)$.

**Bayesian firing probability.** Assume the threshold distribution $q_\phi(\theta)$ is trained variationally to approximate the Bayesian posterior $p(\theta \mid \mathcal{D})$, where $\mathcal{D} = \{(\mathbf{x}_i, y_i)\}_{i=1}^N$ is the training dataset and

$$p(\theta \mid \mathcal{D}) \propto p(\mathcal{D} \mid \theta)\, p(\theta) \quad (38)$$

In the ideal variational limit $q_\phi(\theta) \to p(\theta \mid \mathcal{D})$, the firing probability for a new input $\mathbf{x}$ converges to the Bayesian posterior predictive probability

$$p_{\text{out}}(\mathbf{x}) \to \Pr(S = 1 \mid \mathbf{x}, \mathcal{D}) == \Pr \int (S = 1 \mid \mathbf{x}, \theta)\, p(\theta \mid \mathcal{D})\, d\theta \quad (39)$$

Since $\Pr(S = 1 \mid \mathbf{x}, \theta) = \Pr(D(\mathbf{x}) \geq \theta)$, the spike emission event can be viewed as a Bayesian decision under parameter uncertainty encoded by the posterior over $\theta$.

**A Time Step as A Monte Carlo Sample.** The membrane dynamics of a leaky integrate-and-fire (LIF) neuron can be written as

$$\begin{cases} \text{Charge:} & H_t = [V_{t-1} - \lambda]^+ + X_{\text{in}}(t) \\ \text{Fire:} & S_t = \vartheta(H_t, \theta_t) \\ \text{Reset:} & V_t = V_{t-1}(1 - S_t) \end{cases} \quad (40)$$

where $[z]^+ = \max(0, z))$, $\lambda \geq 0$ is the leak, $X_{\text{in}}(t)$ is the synaptic input at time $t$, and $\theta_t$ is the firing threshold at time $t$.

Inspired by the dynamics of LIF neurons and by the reset behavior of magnetic tunnel junctions (MTJs), we instantiate SB neurons with a memoryless membrane potential. Concretely, we choose a very strong leak such that $\lambda \gg V_{\max}$, where $V_{\max}$ bounds the membrane potential in the operating regime. In this setting, the contribution from the previous state is negligible, so that

$$[V_{t-1} - \lambda]^+ = 0 \quad (41)$$

and the pre-threshold potential simplifies to

$$H_t = X_{\text{in}}(t), \quad \theta_t \overset{\text{i.i.d}}{\sim} q_\phi \quad (42)$$

Under rate encoding, the input $X_{\text{in}}(t)$ is independent across time steps, with statistics determined by $\mathbf{x}$. Together with the i.i.d. thresholds $\theta_t$, this implies that the tuples $(X_{\text{in}}(t), \theta_t)$ and the resulting spikes $S_t$ are independent and identically distributed, with

$$\mathbb{E}[S_t] = p_{\text{out}}(\mathbf{x}) \quad (43)$$

The empirical firing rate over $T$ time steps,

$$\hat{p}_T(\mathbf{x}) = \frac{1}{T}\sum_{t=1}^{T} S_t \tag{44}$$

is therefore an unbiased Monte Carlo estimator of $p_{\text{out}}(\mathbf{x})$, with variance

$$\text{Var}[\hat{p}_T(\mathbf{x})] = \frac{p_{\text{out}}(\mathbf{x})(1 - p_{\text{out}}(\mathbf{x}))}{T} \tag{45}$$

By the strong law of large numbers,

$$\hat{p}_T(\mathbf{x}) \xrightarrow[T\to\infty]{\text{a.s.}} p_{\text{out}}(\mathbf{x}) \tag{46}$$

Each simulation step can thus be interpreted as a Monte Carlo sample from the Spiking Bayesian neuron, and averaging over $T$ steps reduce the estimation variance at the usual $O(T^{-1/2})$ rate.

**Backpropagation and Bayesian Parameter Updates in SBNNs**

Consider a three-layer multilayer perceptron (MLP) implemented as a Spiking Bayesian Neural Network (SBNN). Using the Rate Estimation approximation, the deterministic forward propagation is written as

$$\begin{cases} a_0 = X \\ z_\ell = k\, \dfrac{\mu_{a_{\ell-1}} - \theta_\ell}{\sigma_{a_{\ell-1}}} \\ a_\ell = \sigma(z_\ell), \quad \ell \in \{1,2,3\} \end{cases} \tag{47}$$

where $k \approx 1.716$ is the slope of the logistic approximation, $\sigma(\cdot)$ denotes the logistic sigmoid, and $\mu_{a_{\ell-1}}$ and $\sigma_{a_{\ell-1}}$ are the mean and standard deviation of the input of each neuron at layer $\ell$. These statistics are differentiable functions of the underlying real-valued synaptic parameters and the inputs, obtained after binarisation of the synapses and masking of inactive inputs. The real-valued parameters are projected to quantized weights (1-bit or 8-bit) via a straight-through estimator, so that gradients can be propagated through the binarisation step.

Let $L = L(a_3, Y)$ be the task loss. The local sensitivity at layer $\ell$ is defined as

$$\delta_\ell = \frac{\partial L}{\partial z_\ell} \tag{48}$$

Since

$$a_\ell = \sigma(z_\ell) \tag{49}$$

the chain rule gives

$$\delta_\ell = \frac{\partial L}{\partial a_\ell} \odot a_\ell(1 - a_\ell) \tag{50}$$

where $\odot$ denotes element-wise multiplication.

Using the definition of $z_\ell$, the partial derivatives of the loss with respect to the mean, standard deviation, and threshold terms are

$$\begin{cases} \dfrac{\partial L}{\partial \mu_{a_{\ell-1}}} = \delta_\ell \odot \dfrac{k}{\sigma_{a_{\ell-1}}} \\ \dfrac{\partial L}{\partial \sigma_{a_{\ell-1}}} = \delta_\ell \odot \left(-k\, \dfrac{\mu_{a_{\ell-1}} - \theta_\ell}{\sigma_{a_{\ell-1}}^2}\right) \\ \dfrac{\partial L}{\partial \theta_\ell} = -\sum_b \delta_{\ell,:,b} \odot \dfrac{k}{\sigma_{a_{\ell-1},:,b}} \end{cases} \qquad (51)$$

where the index $b$ runs over the batch, and ":" denotes all units in the layer. For numerical stability, elements with very small $\sigma_{a_{\ell-1}}$ can be treated as effectively deterministic, which are reduced to a hard threshold in the forward path, and the corresponding gradients are suppressed.

The quantities $\mu_{a_{\ell-1}}$ and $\sigma_{a_{\ell-1}}$ are themselves differentiable functions of the synaptic weights and the inputs, typically obtained through linear combinations of masked inputs and simple nonlinear transformations. Gradients with respect to the underlying real-valued weights are then computed automatically by backpropagation through these operations, together with the straight-through approximation used in the binarisation step.

**Threshold Parameterization and Sampling.** Each neuronal threshold is assigned a Gaussian variational posterior. For a given threshold with parameters $\mu$ and $\rho$, the corresponding standard deviation is defined as

$$\sigma_\theta = \log(1 + e^\rho) \qquad (52)$$

and a sample from the posterior is obtained through the reparameterization

$$\tilde{\theta} = \mu + \sigma_\theta \varepsilon, \quad \varepsilon \sim \mathcal{N}(0,1) \qquad (53)$$

To prevent collapse of thresholds toward zero and to avoid excessively excitable neurons, a positive lower bound $\theta_{\min} > 0$ is enforced by clamping,

$$\theta = \max(\tilde{\theta}, \theta_{\min}) \qquad (54)$$

This truncated sampling procedure is used both in the forward pass, where the threshold sample determines the firing rate, and in the backward pass, where gradients are propagated from $\theta$ back to $\mu$ and $\rho$ via the reparameterization.

In practice, several independent samples of $\tilde{\theta}$ are drawn for each neuron. After clamping, the mean of these samples is used as the effective threshold in the forward computation, while the same samples are reused to construct a Monte Carlo estimate of the Kullback–Leibler (KL) divergence between the variational posterior and the prior.

**Bayesian treatment of threshold uncertainty**

Threshold uncertainty is treated in a Bayesian manner using variational inference in the spirit of Bayes-by-Backprop. Let

$$q(\theta \mid \mu, \rho) \qquad (55)$$

denote the approximate posterior over thresholds, with standard deviation Eq.(52). For a dataset $D = \{(x_i, y_i)\}_{i=1}^N$, the variational free energy (negative evidence lower bound) is

$$F(D \mid \mu, \rho) = \mathbb{E}_q[\log q(\theta \mid \mu, \rho)] - \mathbb{E}_q[\log p(\theta)] - \mathbb{E}_q[\log p(D \mid \theta)] \qquad (56)$$

Using the reparameterization as Eq.(53), an unbiased Monte Carlo estimator of the objective with $n$ samples is

$$\hat{F}(D \mid \mu, \rho) = \frac{1}{n} \sum_{i=1}^{n} [\log q(\theta^{(i)} \mid \mu, \rho) - \log p(\theta^{(i)}) - \log p(D \mid \theta^{(i)})] \quad (57)$$

The prior over thresholds is chosen as a two-component Gaussian scale mixture,

$$p(\theta) = \pi \mathcal{N}(\theta; \mu_0, \sigma_1^2) + (1 - \pi) \mathcal{N}(\theta; \mu_0, \sigma_2^2), \quad \sigma_1 > \sigma_2 \quad (58)$$

where $\mu_0$, $\sigma_1$, $\sigma_2$, and $\pi$ are hyperparameters. The prior mean $\mu_0$ is neuron-specific and is initialised by sampling from a normal distribution with prescribed mean and standard deviation. The scale parameters $\sigma_1$ and $\sigma_2$, as well as the mixture weight $\pi$, are kept constant. None of these prior parameters are updated during training, only the variational parameters $\mu$ and $\rho$ are learned.

For a fixed sample of $\theta$, the sample-wise gradient of the Monte Carlo objective with respect to $\theta$ can be written as

$$\frac{\partial \hat{F}}{\partial \theta} = \frac{\partial}{\partial \theta} \log q(\theta \mid \mu, \rho) - \frac{\partial}{\partial \theta} \log p(\theta) - \frac{\partial}{\partial \theta} \log p(D \mid \theta) \quad (59)$$

The likelihood term factorises over data points,

$$\log p(D \mid \theta) = \sum_{j=1}^{N} \log p(y^{(j)} \mid x^{(j)}, \theta) \quad (60)$$

and its gradient with respect to $\theta$ is obtained by backpropagating through the SBNN with the sampled thresholds treated as deterministic parameters for each Monte Carlo draw.

In practice, the KL divergence between $q(\theta \mid \mu, \rho)$ and $p(\theta)$ is evaluated by averaging the difference $\log q(\theta) - \log p(\theta)$ over a fixed number of samples per neuron. This KL term is added to the task loss, and gradients with respect to $\mu$ and $\rho$ are computed by automatic differentiation through the reparameterisation and the truncated sampling procedure. The resulting gradients are then used by a standard stochastic optimizer to update $\mu$ and $\rho$.

**Probabilistic interpretation.** We interpret the spiking process as a Bernoulli random variable obtained by marginalising over the random threshold, and we assume conditional independence of neurons within each layer given their inputs. The stochastic behavior of the SBNN arises at the level of individual neurons. For a neuron $j$ in layer $\ell$, receiving input $x_j^\ell$, the event of emitting a spike $S_j^\ell = 1$ is random. Its probability is determined by the Cumulative Distribution Function (CDF) $F_j^\ell$ of the threshold distribution $p_j^\ell(\theta)$,

$$P(S_j^\ell = 1 \mid S^{\ell-1}) = F_j^\ell(x_j^\ell) = \int_{-\infty}^{x_j^\ell} p_j^\ell(\theta) \, d\theta \quad (61)$$

where $S^{\ell-1}$ is the spike vector in the preceding layer. Assuming conditional independence of thresholds within a layer, the transition probability from layer $\ell - 1$ to layer $\ell$ factorises as

$$P(S^\ell \mid S^{\ell-1}) = \prod_j P(S_j^\ell \mid S^{\ell-1}) \tag{62}$$

with

$$S^\ell = (S_1^\ell, S_2^\ell, \ldots, S_n^\ell) \tag{63}$$

The full network defines a joint distribution over all latent and output spike states, conditioned on the input $S^0$, that factorises along the network topology. The conditional distribution of the output layer is obtained by marginalising over the hidden layers $S^1, \ldots, S^{L-1}$,

$$P(S^L \mid S^0) = \sum_{S^1,\ldots,S^{L-1}} \prod_{\ell=1}^{L} P(S^\ell \mid S^{\ell-1}) \tag{64}$$

Hence, the probability of a particular output spike pattern is the sum of the probabilities over all admissible spike configurations in the hidden layers that can give rise to it.

**Device Characterization and Extraction of Switching Parameters**

To characterize device-level stochastic switching, we measured the antiparallel-to-parallel switching probability of the MTJ as a function of the applied pulse voltage (Table 1 C). For each voltage in the range 0.38–0.60 V with a step size of 0.01 V, 10,000 identical pulses were applied. The MTJ state after each pulse was determined from the post-pulse resistance readout, and the switching probability was computed as the fraction of pulses that induced a successful transition. The resulting probability–voltage relation was fit using a sigmoid function, yielding a characteristic switching voltage $V_{50} = 0.4997 \pm 0.014$V (mean $\pm$s.d.), defined as the voltage corresponding to 50% switching probability.

**Hardware Measurement and Inference Setup**

To ensure compatibility between the algorithmic SBNN and the MTJ hardware, we rescaled the neuronal inputs during inference so that each neuron's effective threshold aligned with the measured device switching characteristic. For a neuron with learned threshold $\mu$ and algorithmic input $x_{\text{in}}$, the modulated input was defined as

$$x_{\text{mod}} = \frac{V_{50} \cdot x_{\text{in}}}{\mu} \tag{65}$$

where $V_{50} = 0.4997$V is obtained from the sigmoid fit described above. The modulated inputs were subsequently mapped to MTJ drive voltages according to this expression, enabling neuron-specific threshold alignment while using a common device switching curve.

All electrical measurements were performed using a Keithley 4200A-SCS semiconductor parameter analyzer equipped with a 4225-PMU pulse measure unit and a 4225-RPM remote amplifier/switch.

**Device Inference Procedure and Readout**

Hardware inference was performed by applying the modulated inputs to MTJ and reading out their post-pulse resistance states at each time step. Specifically, the hidden-layer activity was realized by (i) converting the modulated inputs at each time step into MTJ drive voltages, (ii) applying the corresponding voltage pulses to MTJ, and (iii) determining the MTJ state after each pulse from resistance readout, thereby producing hidden-layer spike trains over 16 steps. These hidden-layer spike trains were then linearly combined using the learned synaptic weights to compute the inputs to the output layer. Output neurons were implemented using the same MTJ-based procedure, and their spiking rates accumulated over the 16 time steps formed the final network readout. The predicted class label was obtained from the output-layer activity, by selecting the class with the largest output firing rate.

# Supplementary Materials

## Supplementary Text

**Note 1:** Experiment Setting of Image Recognition Tasks

We evaluate our spiking Bayesian neural network (SBNN) on MNIST, Fashion MNIST and Cifar10. Inputs are normalized to $[0,1]$ and flattened to vectors.

**Note 2:** Encoding of the Inputs

For all inputs, we used a Poisson encoder to convert normalized intensities $x \in [0,1]$ into spike trains of length $T$. For each element (pixel/feature) and each time step $t$, we generated a spike by drawing $u_t \sim \text{Uniform}(0,1)$ and setting
$$s_t = \mathbb{I}[u_t < x],$$
which is equivalent to $s_t \sim \text{Bernoulli}(x)$. This process is performed independently across time steps and input dimensions, yielding outputs of shape $(B, \ldots, T)$.

**Fig. S1.**

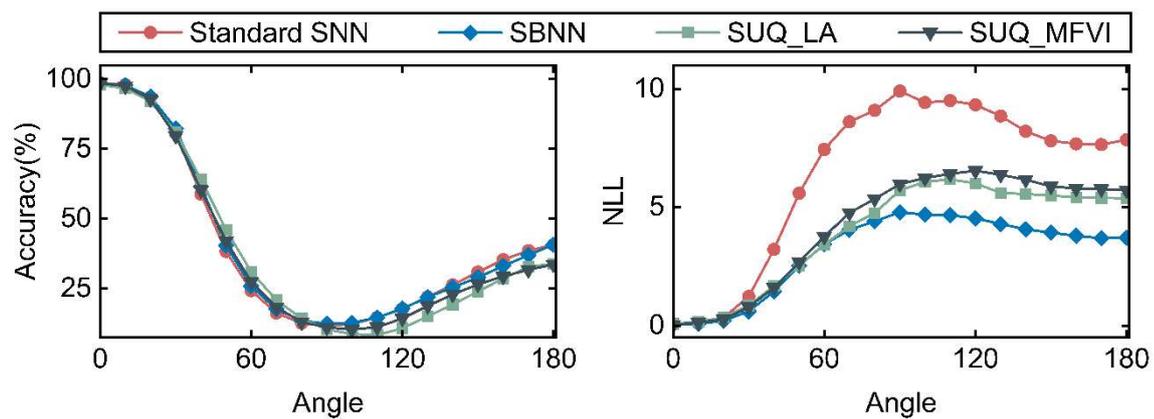

Figure S1: for MNIST-trained MLP on rotated versions of the MNIST test set. The rotation degree interval is 10 from 0 – 180. The comparison models (SUQ_LA and SUQ_MFVI) from the article [1].

**Fig. S2.**

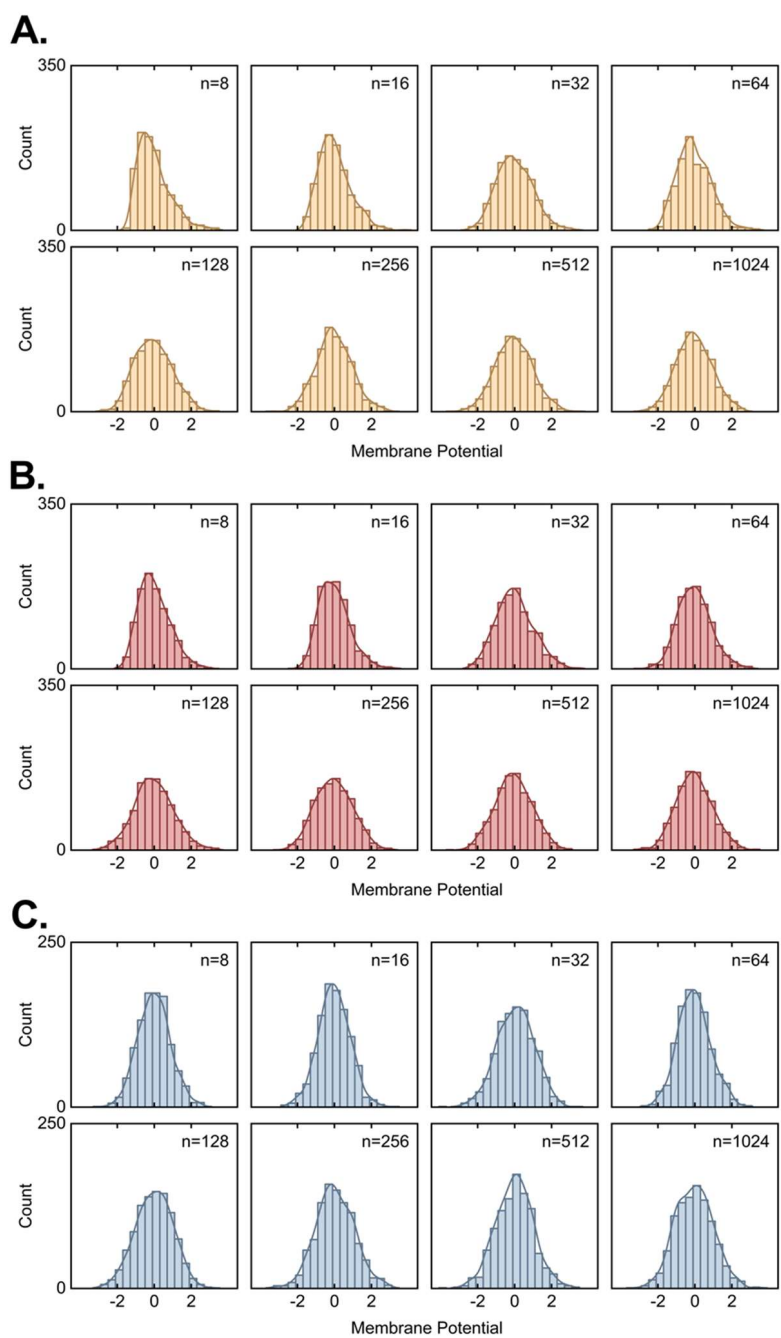

Figure S2: The distribution of membrane potential, where the uncertainty of the weights belongs to the exponential family distribution. (A) Exponential. (B) Gamma. (C) Gaussian.

**Fig. S3.**

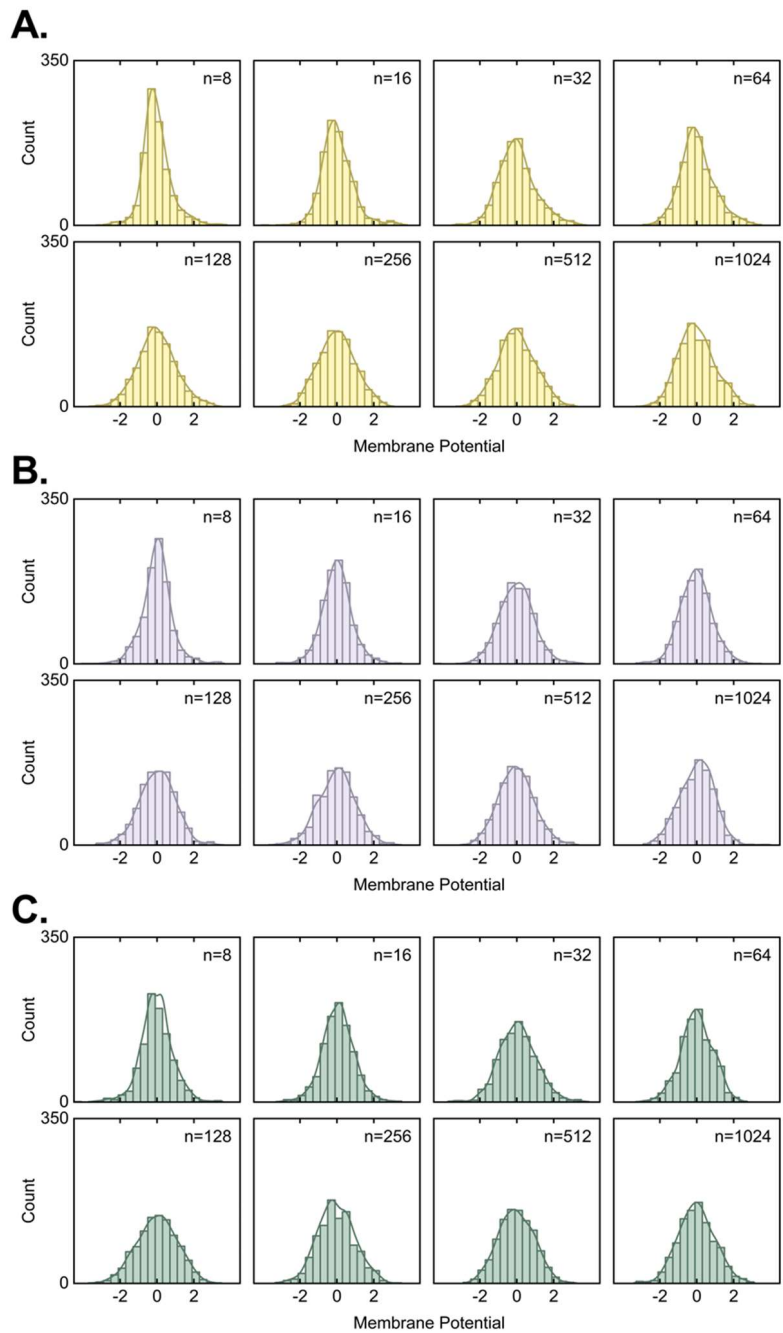

Figure S3: The distribution of membrane potential, where the uncertainty of the weights belongs to the non-exponential family distribution. (A) Lévy Stable. (B) Student-T. (C) Symmetric Pareto.

**Fig. S4.**

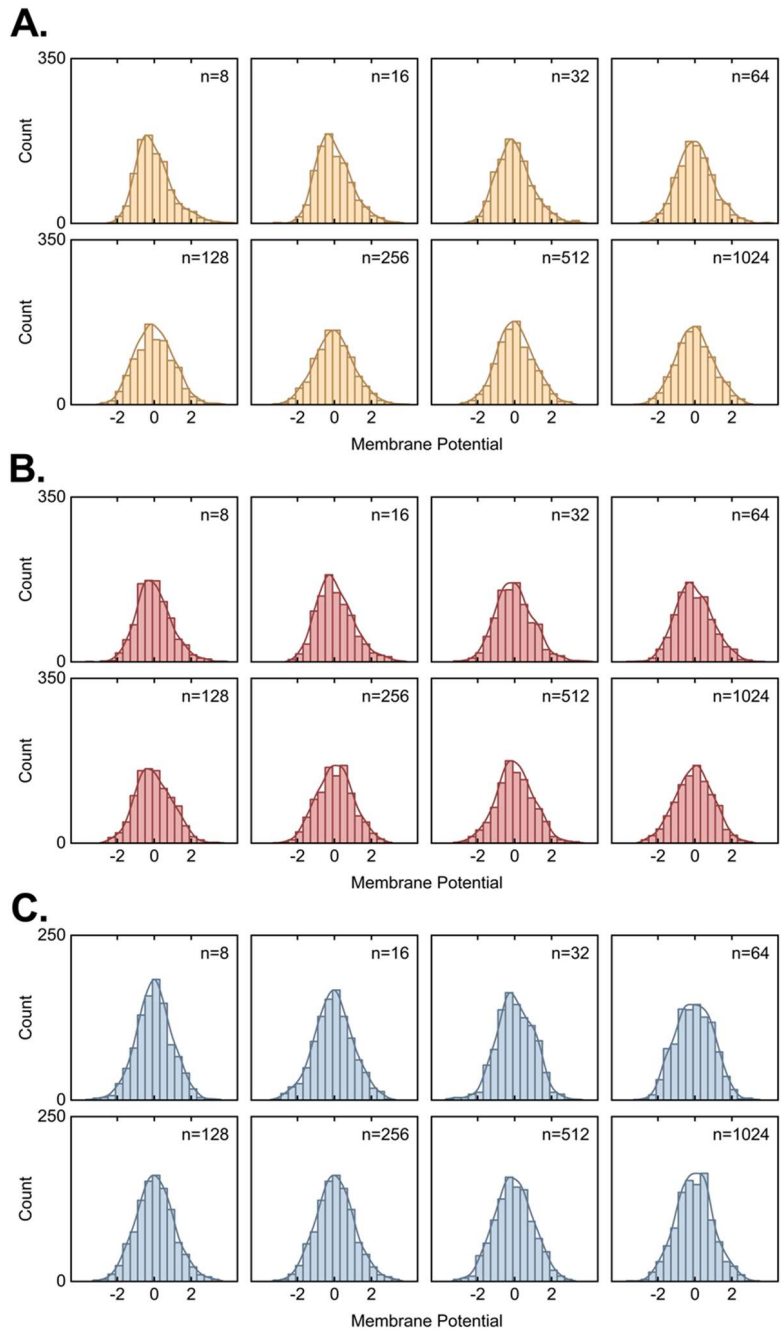

Figure S4: The distribution of membrane potential, where the uncertainty of the weights belongs to the exponential family distribution. In addition, the inputs are random in different samples. (A) Exponential. (B) Gamma. (C) Gaussian.

**Fig. S5.**

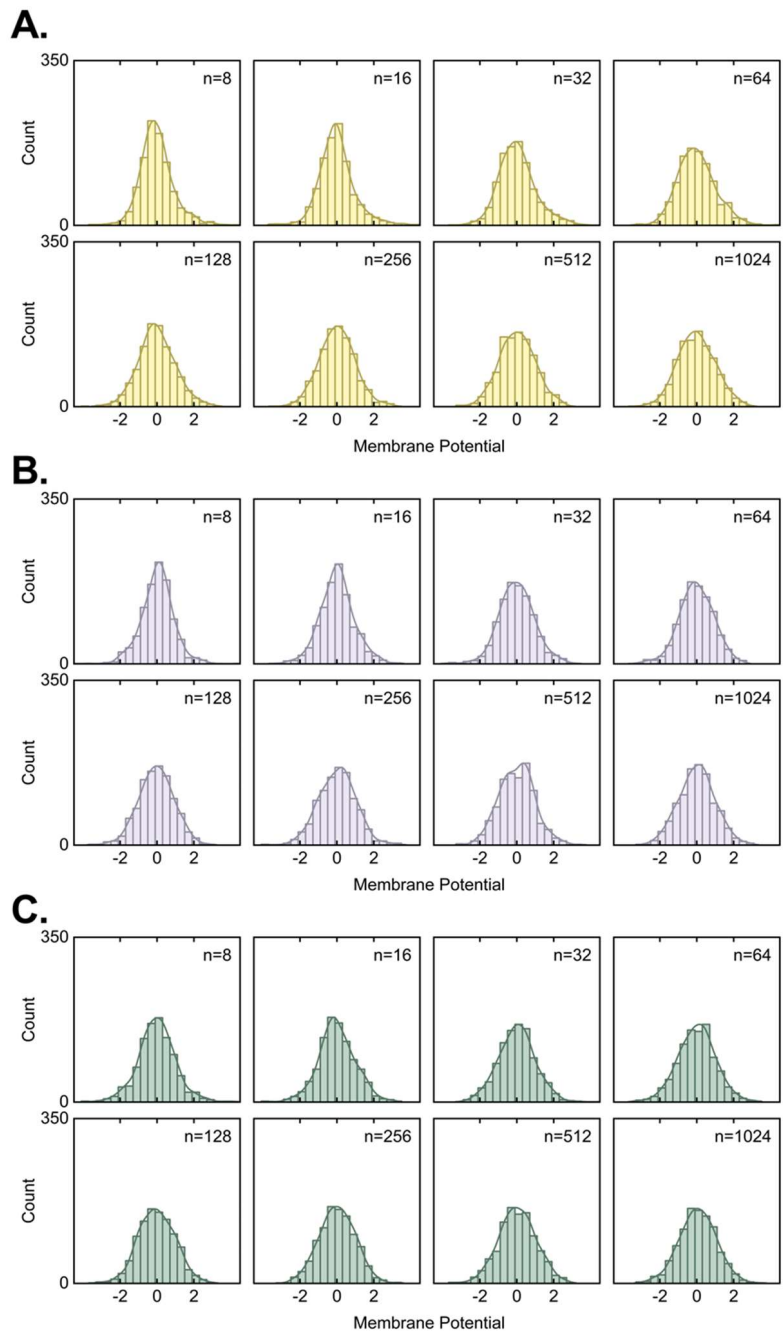

Figure S5: The distribution of membrane potential, where the uncertainty of the weights belongs to the non-exponential family distribution. In addition, the inputs are random in different samples. (A) Lévy Stable. (B) Student-T. (C) Symmetric Pareto.

**Fig. S6.**

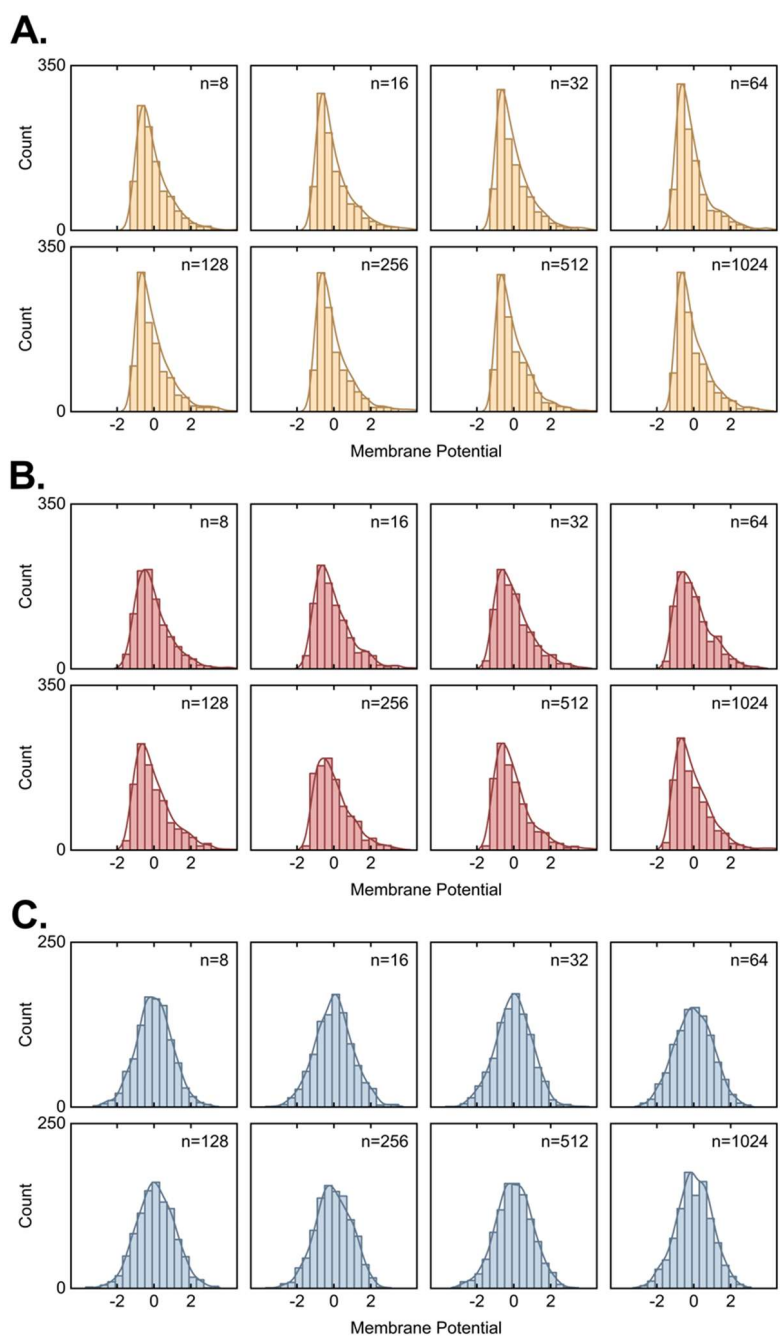

Figure S6: The distribution of membrane potential, where the uncertainty of the neurons belongs to the exponential family distribution. (A) Exponential. (B) Gamma. (C) Gaussian.

**Fig. S7.**

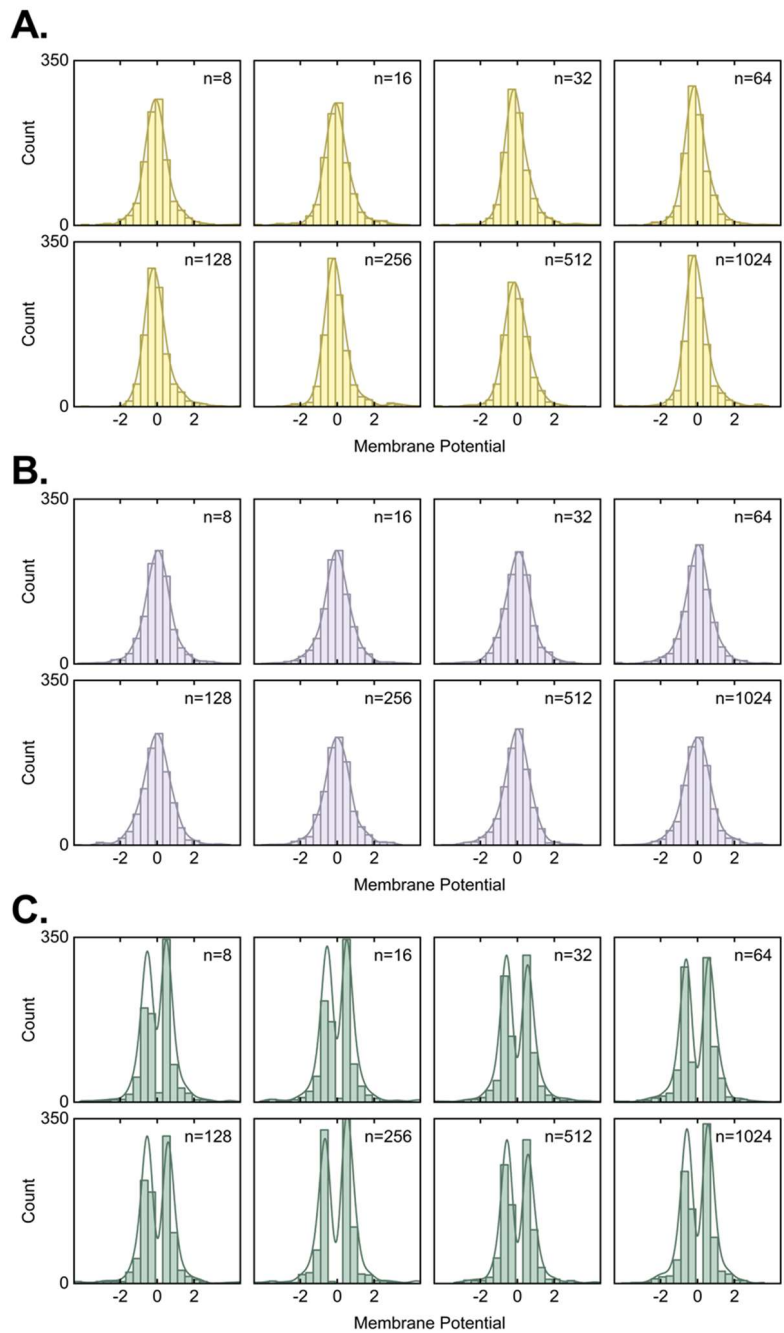

Figure S7: The distribution of membrane potential, where the uncertainty of the neurons belongs to the non-exponential family distribution. (A) Lévy Stable. (B) Student-T. (C) Symmetric Pareto.

**Fig. S8.**

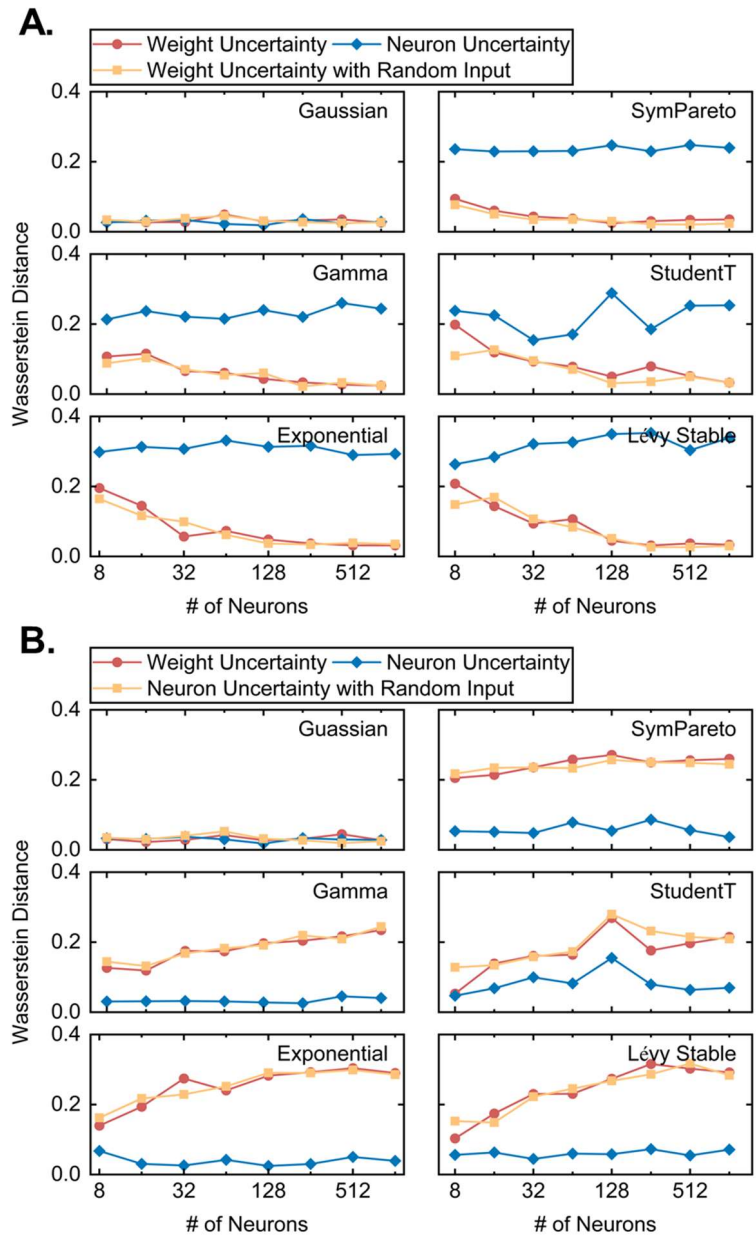

Figure S8: Across various number of neurons, the Wasserstein-1 distance between the distribution of membrane potential and the standard distribution or standard Gaussian distribution. (A) standard Gaussian distribution. (B) standard distribution.

**Fig. S9.**

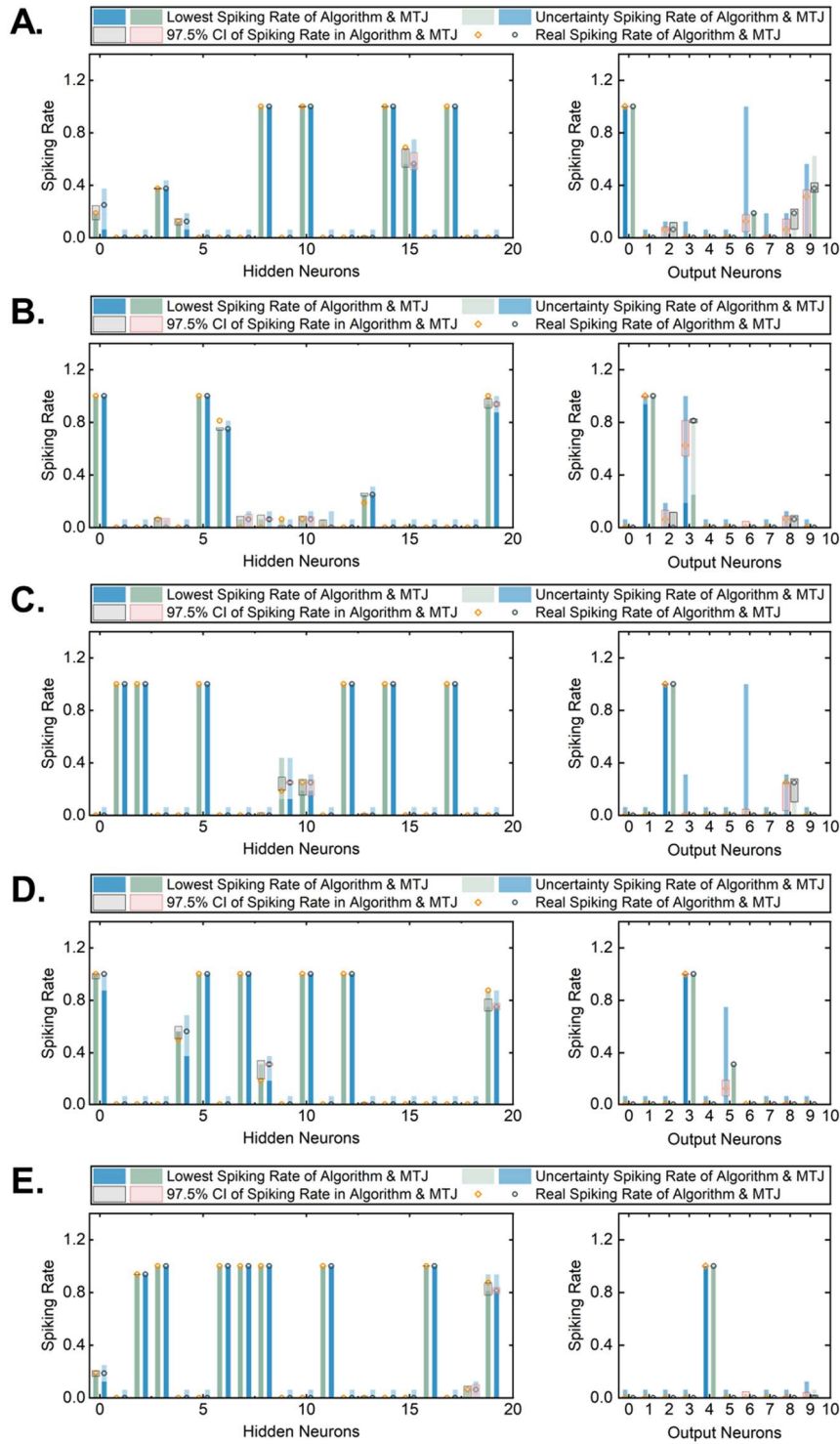

Figure S9: Hidden layer and Output layer spiking rates for MNIST test set predicted by the algorithm and measured on MTJ hardware. (A) Class 0, Sample 1565. (B) Class 1, Sample 191. (C) Class 2, Sample 298. (D) Class 3, Sample 2639. (E) Class 4, Sample 3792.

**Fig. S10.**

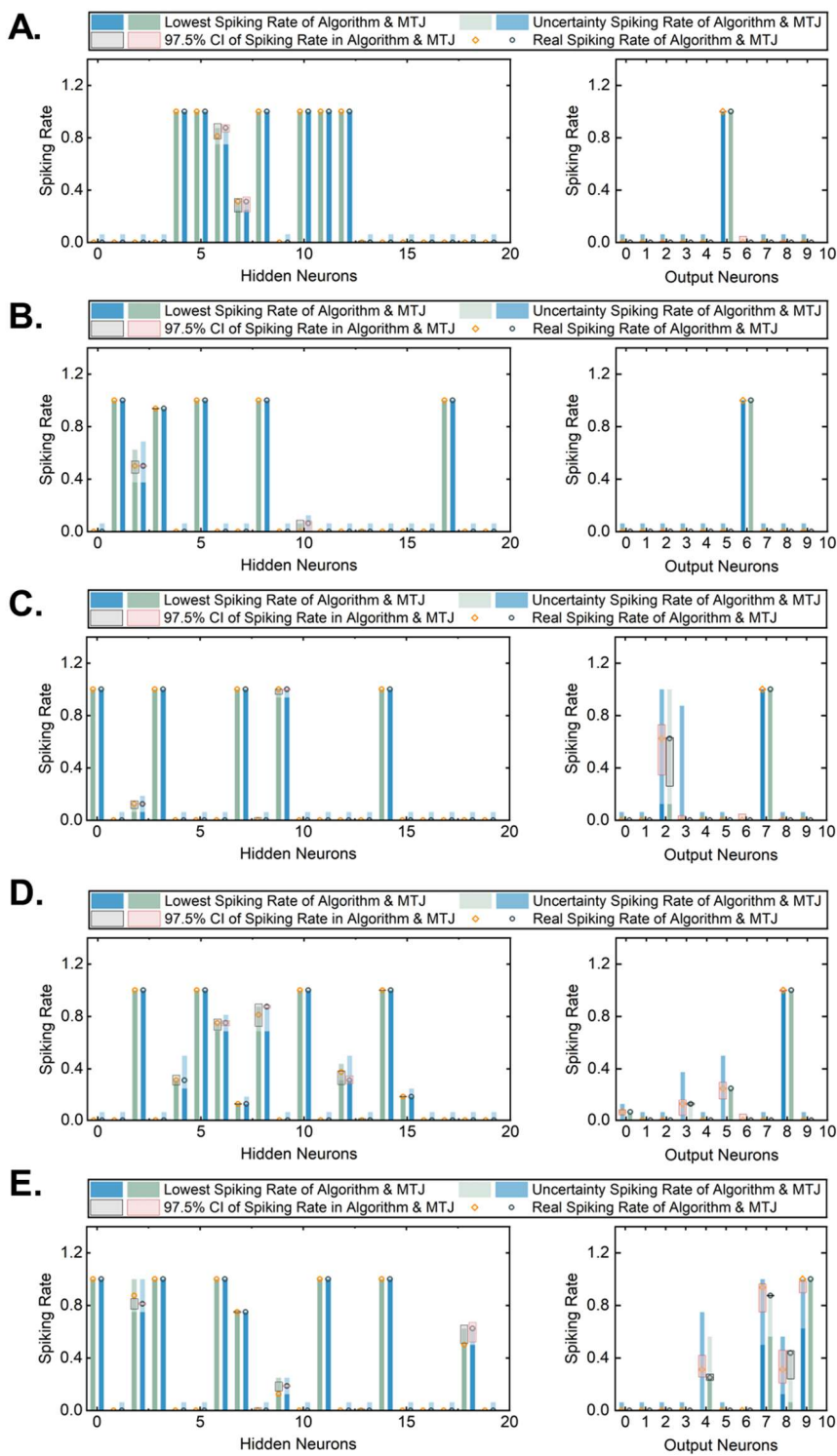

Figure S10: Hidden layer and Output layer spiking rates for MNIST test set predicted by the algorithm and measured on MTJ hardware. (A) Class 5, Sample 1565. (B) Class 6, Sample 191. (C) Class 7, Sample 298. (D) Class 8, Sample 2639. (E) Class 9, Sample 3792.

**Fig. S11.**

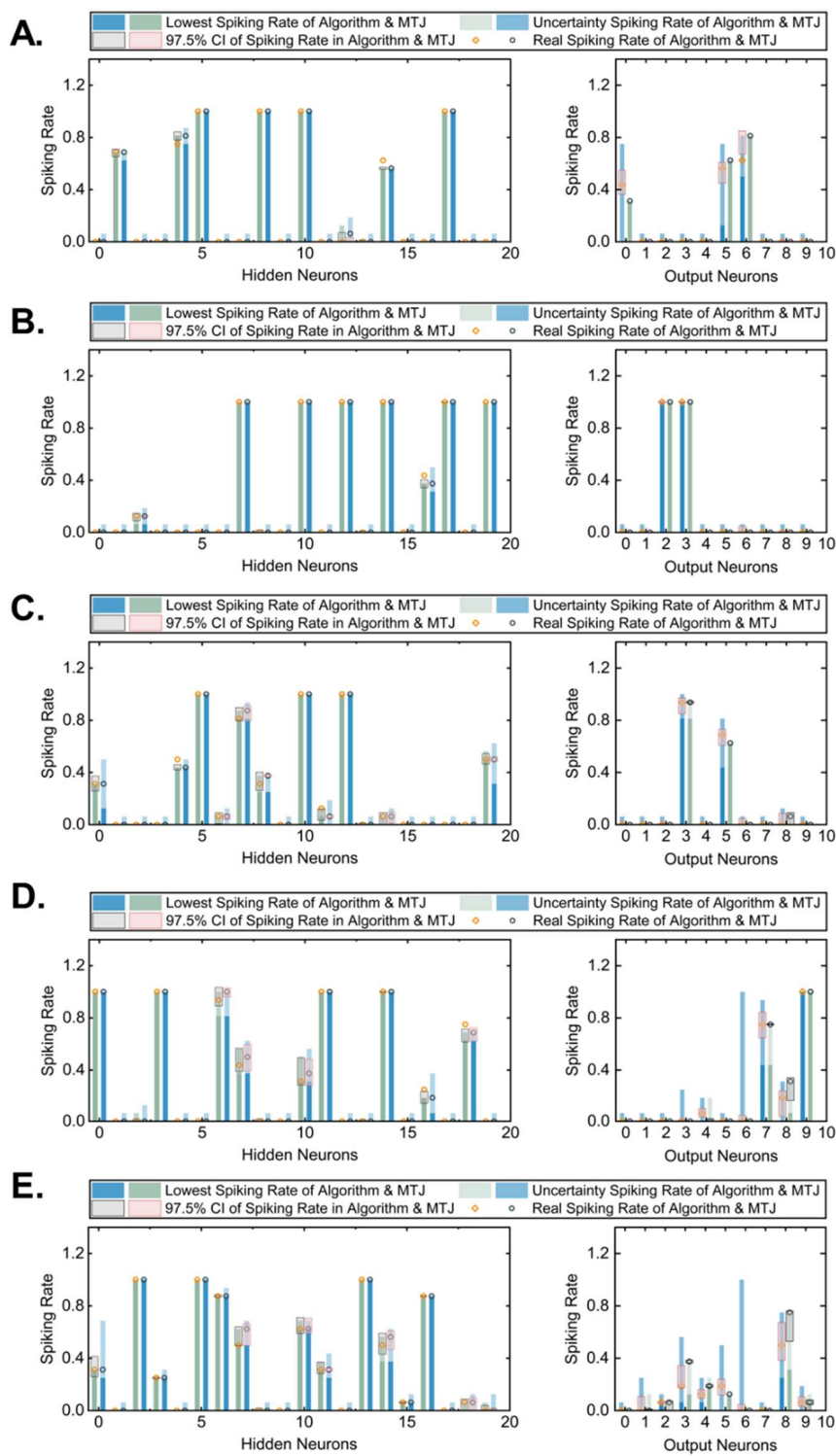

Figure S11: Hidden layer and Output layer spiking rates for MNIST test set predicted by the algorithm and measured on MTJ hardware (Error Samples). (A) Class 0, Sample 5255. (B) Class 3, Sample 4833. (C) Class 5, Sample 2832. (D) Class 7, Sample 5600. (E) Class 9, Sample 4761.

**Fig. S12.**

| | Samples | | | | | | | | | |
|---|---|---|---|---|---|---|---|---|---|---|
| 0 | 1565 | 2733 | 4079 | 5255 | 5929 | 6286 | 6400 | 6489 | 6752 | 9601 |
| 1 | 191 | 1038 | 3454 | 4179 | 4674 | 8491 | 8526 | 9795 | 9923 | 9950 |
| 2 | 298 | 816 | 1177 | 2940 | 2971 | 3128 | 4504 | 7637 | 7785 | 9942 |
| 3 | 349 | 2639 | 2952 | 4833 | 6107 | 7312 | 7968 | 8393 | 8811 | 9342 |
| 4 | 3718 | 3792 | 4266 | 5926 | 7341 | 7456 | 8193 | 8312 | 9099 | 9605 |
| 5 | 356 | 1115 | 1510 | 1525 | 2518 | 2832 | 4054 | 4979 | 5598 | 7630 |
| 6 | 3331 | 4239 | 4622 | 4814 | 5599 | 6258 | 6842 | 8423 | 8990 | 9149 |
| 7 | 617 | 2091 | 3225 | 4214 | 4693 | 5600 | 6589 | 6762 | 7069 | 8248 |
| 8 | 714 | 1371 | 2859 | 3064 | 5049 | 6001 | 6617 | 6654 | 8699 | 8934 |
| 9 | 193 | 639 | 962 | 1597 | 3041 | 3723 | 4237 | 4761 | 8002 | 8998 |

Figure S12: All samples of MNIST dataset in our hardware experiments.

**Fig. S13.**

**A.**

**Algorithm 1:** QTsize8Weight (8-bit quantization + STE)

**Data:** $x$ (2D tensor), $Con \in \mathbb{R}^{C_{out} \times C_{in}}$, sigmaType
**Result:** $\mu \in \mathbb{R}^{C_{out} \times B}$, $\sigma \in \mathbb{R}^{C_{out} \times B}$

$w_{max} \leftarrow 1.0$  step $\leftarrow (2 \cdot w_{max})/255$
$w \leftarrow \text{Clamp}(Con, -w_{max}, w_{max})$
$W \leftarrow \text{Round}((w + w_{max})/step) \cdot step - w_{max}$
$C_{in} \leftarrow \text{cols}(W)$
**if** $rows(x) == C_{in}$ **then**
 $\quad X \leftarrow x$;  // $x$ is $C_{in} \times B$
**else**
 $\quad$ **if** $cols(x) == C_{in}$ **then**
 $\quad\quad X \leftarrow x^T$;  // $x$ is $B \times C_{in}$
 $\quad$ **else**
 $\quad\quad$ error;  // feature mismatch
 $\quad$ **end**
**end**
$mask \leftarrow 1[X > 1/512]$
$X_{mask} \leftarrow X \odot mask$
$\mu \leftarrow W \cdot X_{mask}$
$v \leftarrow X_{mask} \odot (1 - X_{mask})$
**if** $sigmaType == sq$ **then**
 $\quad \sigma^2 \leftarrow (W \odot W) \cdot v$
**else**
 $\quad \sigma^2 \leftarrow |W| \cdot v$
**end**
$\sigma \leftarrow \sqrt{\max(\sigma^2, 10^{-20})}$
**return** $(\mu, \sigma)$

**Parameters**

**x**
 indicates the inputs
**mu**
 indicates the mean of input spikes
**sigma**
 indicates the standard deviation of input spikes
**threshold**
 indicates the threshold of corresponded neurons
**k**
 indicates the parameter of probability estimate
**eps**
 indicates a extremely small constant

**B.**

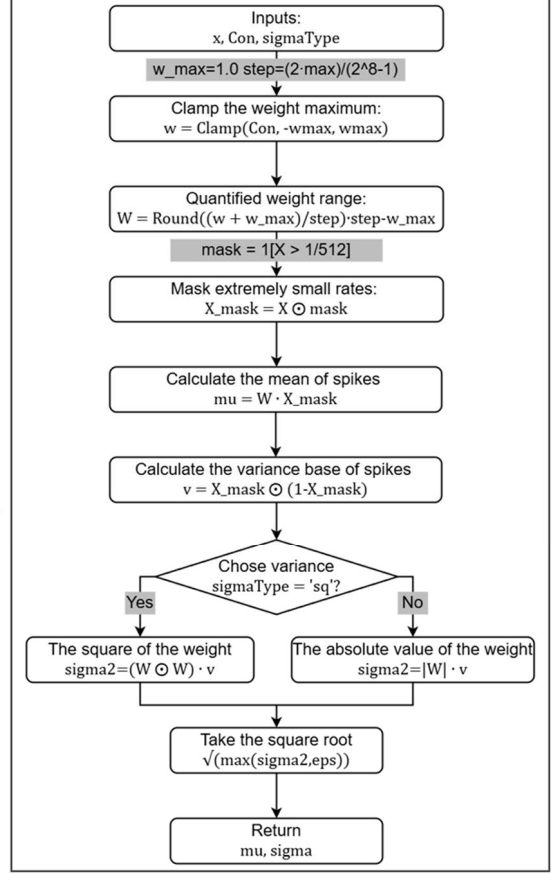

Figure S13: Forward and backward passes for weight quantization. (A) Pseudocode for QTsize8Weight. (B) The encapsulated QTsize8Weight, used for weight quantization and calculate the mean and standard deviation of input spikes for neurons in the linear layer.

**Fig. S14.**

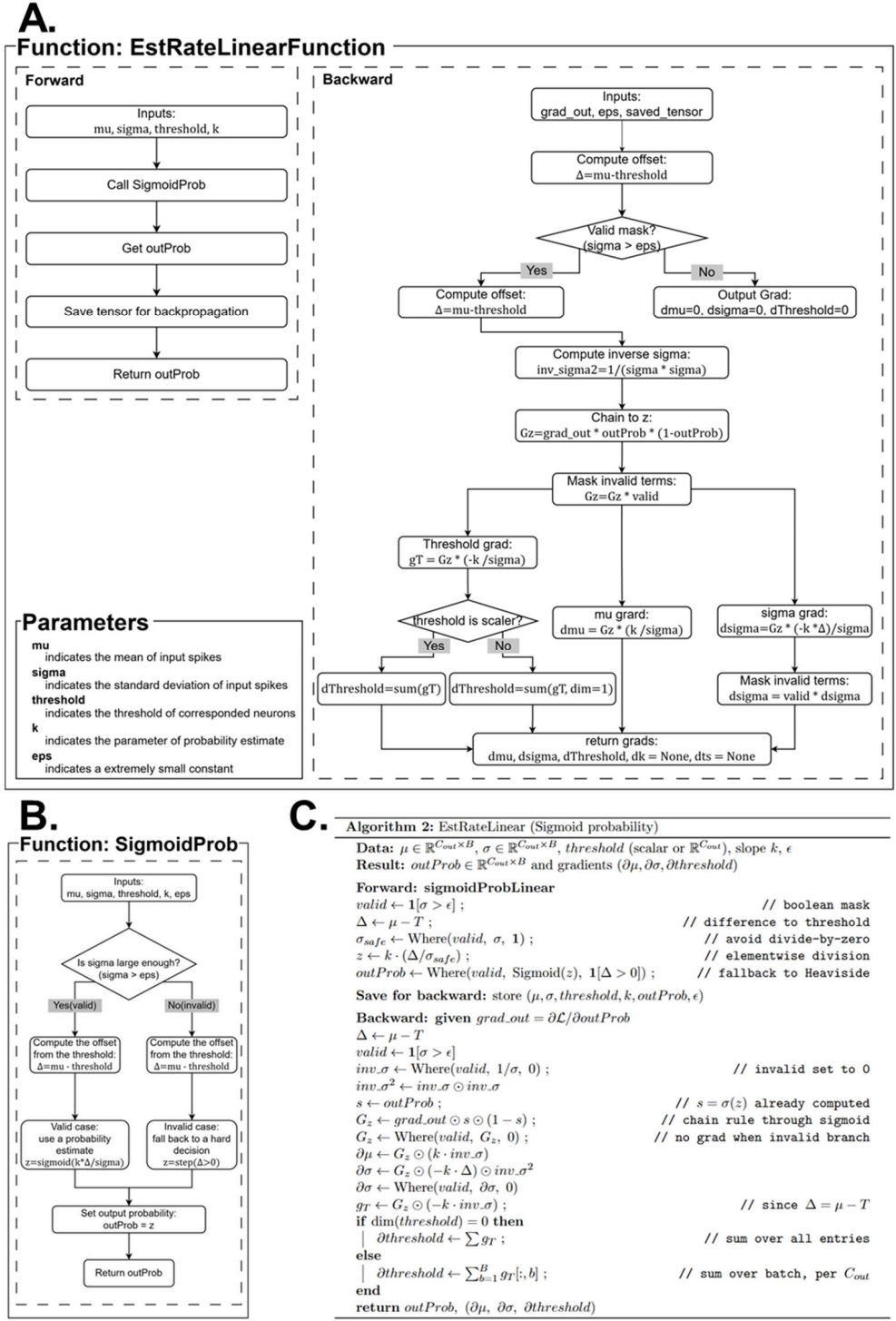
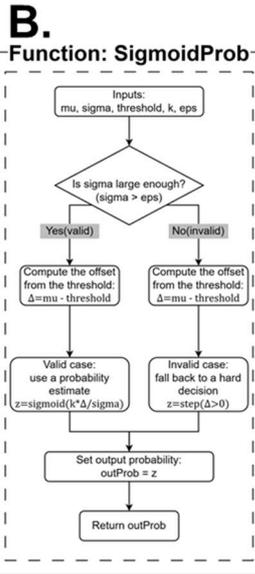

Figure S14: Forward and backward passes for rate estimation. (A) The encapsulated EstRateLinearFunction, used for rate estimation in the linear layer. (B) The main function for rate estimation, SigmoidProb. (C) Pseudocode for rate estimation.

**Fig. S15.**

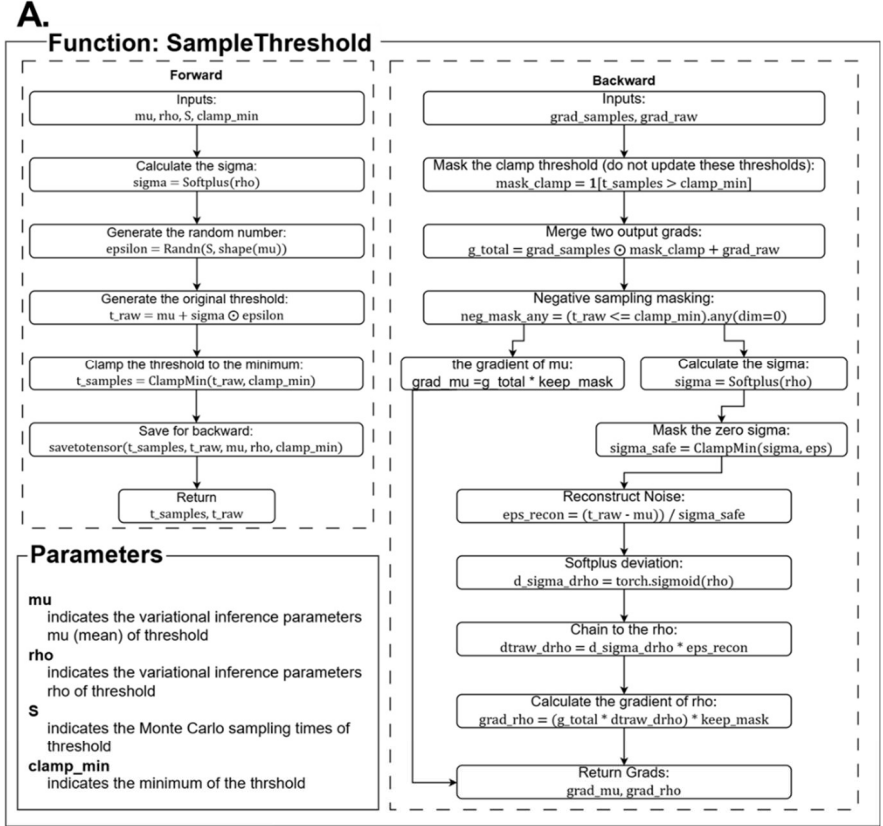

Figure S15: Forward and backward passes for SampleThreshold. (A) The encapsulated SampleThreshold, used for sampling the threshold at each forward. (B) Pseudocode for SampleThreshold.

Fig. S16.

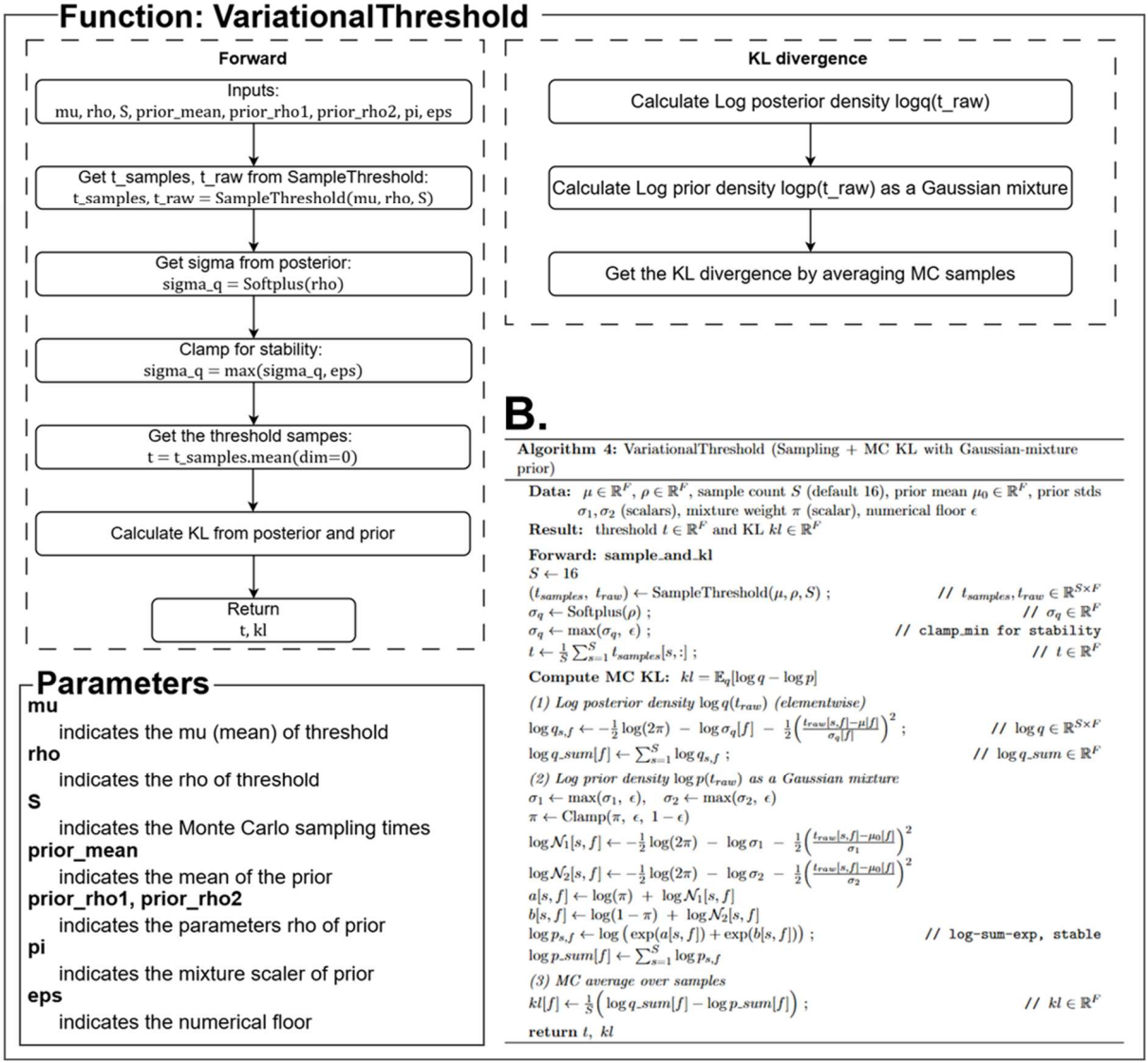

Figure S16: Forward passes for VariationalThreshold. (A) The encapsulated VariationalThreshold, used for sampling the threshold and calculating the KL divergence at each forward. (B) Pseudocode for VariationalThreshold.

# S3. Supplementary Table

Table S1: Accuracy of SBNN in Recognition Tasks (%)

| Data Set | Weight bits | Rho | Time Steps | | | | | |
|---|---|---|---|---|---|---|---|---|
| | | | 2 | 3 | 4 | 6 | 8 | 16 |
| MNIST | 8 | -0.5 | 98.77 | 98.90 | 99.04 | 99.09 | 99.16 | 99.15 |
| | | 0.0 | 98.80 | 98.88 | 98.99 | 99.08 | 99.14 | 99.16 |
| | | 0.5 | 98.79 | 98.99 | 99.02 | 99.09 | 99.18 | 99.14 |
| | 1 | -0.5 | 97.89 | 98.12 | 98.32 | 98.51 | 98.56 | 98.71 |
| | | 0.0 | 97.75 | 98.15 | 98.35 | 98.37 | 98.53 | 98.57 |
| | | 0.5 | 97.72 | 98.22 | 98.30 | 98.43 | 98.54 | 98.62 |
| Fashion MNIST | 8 | -0.5 | 86.38 | 87.79 | 88.20 | 89.20 | 89.44 | 90.05 |
| | | 0.0 | 86.79 | 87.50 | 88.30 | 88.98 | 89.48 | 89.95 |
| | | 0.5 | 85.92 | 87.70 | 88.38 | 89.00 | 89.45 | 89.84 |
| | 1 | -0.5 | 82.67 | 84.47 | 84.79 | 85.66 | 86.20 | 86.71 |
| | | 0.0 | 83.48 | 84.57 | 85.94 | 86.39 | 86.49 | 87.31 |
| | | 0.5 | 83.27 | 84.43 | 85.78 | 86.33 | 86.53 | 87.16 |
| Cifar10 | Mix | -0.5 | 94.37 | 94.41 | 94.22 | 94.37 | 94.29 | 94.58 |
| | | 0.0 | 94.23 | 94.26 | 94.16 | 94.35 | 94.62 | 94.39 |
| | | 0.5 | 94.64 | 94.39 | 94.61 | 94.65 | 94.84 | 94.70 |

Table S2: Experiment Setting of Image Recognition Tasks

| | MNIST | F-MNIST | Cifar10 |
|---|---|---|---|
| Optimizer | adamw | | |
| Batch Size | 64 | 64 | 64 |
| Epochs | 500 | 2000 | 1000 |
| Sigmoid Scale Factor (K) | 1.716 | 1.716 | 1.716 |
| Threshold Initialization | 1.0 | 1.0 | 2.0 |
| Minium of Threshold | 1/128 | | |
| Learning Rate (Threshold) | 1e-5 | 7.5e-4 | 1e-5 |
| Weight Quantization Range | [-1, 1] | | |
| Learning Rate (Weight) | 5e-4 | 5e-4 | 5e-3 |
| KL Beta | 1e-3 | 1e-5 | 5e-8 |
| Loss Function | Cross-Entropy | | |
| Learning Rate Scheduler | Cosine Annealing LR | | |
| Weight Decay | 1e-3 | | |
| Scheduler Max Epochs | 500 | 1000 | 1000 |
| Prior Mu Initialization (Mean) | 3.0 | 3.0 | 3.0 |
| Prior Mu Initialization (Std) | 2.9 | 1.5 | 1.5 |
| Std of Prior1 | 0.5 | 1.625 | 0.5 |
| Std of Prior2 | 0.05 | 0.05 | 0.05 |
| Mixture Ratio of Prior | 0.5 | | |
| Seed | 42 | | |
| Random Crop | (28, 1) | | (32, 4) |
| Random Horizontal Flip | False | | True |
| Cutout | False | | 1/16 |
| Warm Up Epochs | 0 | | 20 |

Table S3: Device Switching Probability. For all sweeping voltages, we applied 10,000 independent times to the device and read out the device resistance after applying voltages.

| Voltage (V) | Parallel (# of Pulse) | Anti-Parallel (# of Pulse) | Switching Probability |
|---|---|---|---|
| 0.6 | 10000 | 0 | 1 |
| 0.59 | 10000 | 0 | 1 |
| 0.58 | 9999 | 1 | 0.9999 |
| 0.57 | 9979 | 21 | 0.9979 |
| 0.56 | 9903 | 97 | 0.9903 |
| 0.55 | 9711 | 289 | 0.9711 |
| 0.54 | 9329 | 671 | 0.9329 |
| 0.53 | 8634 | 1366 | 0.8634 |
| 0.52 | 7613 | 2387 | 0.7613 |
| 0.51 | 6373 | 3627 | 0.6373 |
| 0.50 | 4899 | 5101 | 0.4899 |
| 0.49 | 3674 | 6326 | 0.3674 |
| 0.48 | 2642 | 7358 | 0.2642 |
| 0.47 | 1750 | 8250 | 0.175 |
| 0.46 | 1296 | 8704 | 0.1296 |
| 0.45 | 805 | 9195 | 0.0805 |
| 0.44 | 501 | 9499 | 0.0501 |
| 0.43 | 310 | 9690 | 0.031 |
| 0.42 | 190 | 9810 | 0.019 |
| 0.41 | 106 | 9894 | 0.0106 |
| 0.4 | 42 | 9958 | 0.0042 |
| 0.39 | 28 | 9972 | 0.0028 |
| 0.38 | 9 | 9991 | 0.0009 |
| 0.37 | 1 | 9999 | 0.0001 |

Table S4: Spiking Rate of Each Neuron in Algorithmic Model and Device Implementation (Class 0 & Class 1)

| Class | Type | Sample | Neuron | | | | | | | | | |
|---|---|---|---|---|---|---|---|---|---|---|---|---|
| | | | 0 | 1 | 2 | 3 | 4 | 5 | 6 | 7 | 8 | 9 |
| 0 | Algorithm | 1565 | **1.00** | 0.00 | 0.06 | 0.00 | 0.00 | 0.00 | 0.13 | 0.00 | 0.06 | 0.31 |
| | | 2733 | **1.00** | 0.00 | 0.00 | 0.00 | 0.00 | 0.00 | 0.13 | 0.00 | 0.13 | 0.06 |
| | | 4079 | **0.94** | 0.00 | 0.00 | 0.00 | 0.00 | 0.00 | 0.00 | 0.00 | 0.06 | 0.00 |
| | | 5255 | 0.44 | 0.00 | 0.00 | 0.00 | 0.00 | 0.56 | **0.63** | 0.00 | 0.00 | 0.00 |
| | | 5929 | **1.00** | 0.00 | 0.06 | 0.00 | 0.00 | 0.00 | 0.00 | 0.00 | 0.06 | 0.00 |
| | | 6286 | **1.00** | 0.00 | 0.00 | 0.00 | 0.00 | 0.00 | 0.00 | 0.00 | 0.00 | 0.00 |
| | | 6400 | **1.00** | 0.00 | 0.00 | 0.00 | 0.00 | 0.00 | 0.88 | 0.00 | 0.00 | 0.00 |
| | | 6489 | **1.00** | 0.00 | 0.00 | 0.00 | 0.00 | 0.19 | 0.00 | 0.00 | 0.00 | 0.00 |
| | | 6752 | **1.00** | 0.00 | 0.00 | 0.00 | 0.00 | 0.00 | 0.00 | 0.00 | 0.00 | 0.00 |
| | | 9601 | **1.00** | 0.00 | 0.00 | 0.00 | 0.00 | 0.00 | 0.00 | 0.00 | 0.00 | 0.00 |
| | Device | 1565 | **1.00** | 0.00 | 0.06 | 0.00 | 0.00 | 0.00 | 0.19 | 0.00 | 0.19 | 0.38 |
| | | 2733 | **1.00** | 0.00 | 0.00 | 0.00 | 0.00 | 0.00 | 0.13 | 0.00 | 0.13 | 0.13 |
| | | 4079 | **1.00** | 0.00 | 0.00 | 0.00 | 0.00 | 0.00 | 0.00 | 0.00 | 0.00 | 0.00 |
| | | 5255 | 0.31 | 0.00 | 0.00 | 0.00 | 0.00 | 0.63 | **0.81** | 0.00 | 0.00 | 0.00 |
| | | 5929 | **1.00** | 0.00 | 0.00 | 0.00 | 0.00 | 0.00 | 0.00 | 0.00 | 0.06 | 0.00 |
| | | 6286 | **1.00** | 0.00 | 0.00 | 0.00 | 0.00 | 0.00 | 0.00 | 0.00 | 0.00 | 0.00 |
| | | 6400 | **0.94** | 0.00 | 0.00 | 0.00 | 0.00 | 0.00 | 1.00 | 0.00 | 0.00 | 0.00 |
| | | 6489 | **1.00** | 0.00 | 0.00 | 0.00 | 0.00 | 0.19 | 0.00 | 0.00 | 0.00 | 0.00 |
| | | 6752 | **1.00** | 0.00 | 0.00 | 0.00 | 0.00 | 0.00 | 0.00 | 0.00 | 0.00 | 0.00 |
| | | 9601 | **1.00** | 0.00 | 0.00 | 0.00 | 0.00 | 0.00 | 0.00 | 0.00 | 0.00 | 0.00 |
| 1 | Algorithm | 0191 | 0.00 | **1.00** | 0.06 | 0.63 | 0.00 | 0.00 | 0.00 | 0.00 | 0.06 | 0.00 |
| | | 1038 | 0.00 | **1.00** | 0.00 | 0.13 | 0.00 | 0.13 | 0.00 | 0.00 | 0.00 | 0.00 |
| | | 3454 | 0.00 | **1.00** | 0.00 | 0.00 | 0.00 | 0.00 | 0.00 | 0.00 | 0.00 | 0.00 |
| | | 4179 | 0.00 | **1.00** | 0.00 | 0.00 | 0.00 | 0.00 | 0.00 | 0.00 | 0.00 | 0.00 |
| | | 4674 | 0.00 | **1.00** | 0.06 | 0.00 | 0.00 | 0.00 | 0.00 | 0.00 | 0.00 | 0.00 |
| | | 8491 | 0.00 | **1.00** | 0.00 | 0.00 | 0.00 | 0.00 | 0.00 | 0.00 | 0.00 | 0.00 |
| | | 8526 | 0.00 | **1.00** | 0.00 | 0.00 | 0.00 | 0.00 | 0.00 | 0.00 | 0.00 | 0.00 |
| | | 9795 | 0.00 | **1.00** | 0.00 | 0.00 | 0.00 | 0.00 | 0.00 | 0.00 | 0.31 | 0.00 |
| | | 9926 | 0.00 | **1.00** | 0.00 | 0.00 | 0.00 | 0.00 | 0.00 | 0.00 | 0.00 | 0.00 |
| | | 9950 | 0.00 | **1.00** | 0.00 | 0.00 | 0.00 | 0.00 | 0.00 | 0.00 | 0.19 | 0.00 |
| | Device | 0191 | 0.00 | **1.00** | 0.00 | 0.81 | 0.00 | 0.00 | 0.00 | 0.00 | 0.06 | 0.00 |
| | | 1038 | 0.00 | **1.00** | 0.00 | 0.13 | 0.00 | 0.13 | 0.00 | 0.00 | 0.00 | 0.00 |
| | | 3454 | 0.00 | **1.00** | 0.00 | 0.00 | 0.00 | 0.00 | 0.00 | 0.00 | 0.00 | 0.00 |
| | | 4179 | 0.00 | **1.00** | 0.06 | 0.00 | 0.00 | 0.00 | 0.00 | 0.00 | 0.00 | 0.00 |
| | | 4674 | 0.00 | **1.00** | 0.13 | 0.00 | 0.00 | 0.00 | 0.00 | 0.00 | 0.00 | 0.00 |
| | | 8491 | 0.00 | **1.00** | 0.06 | 0.00 | 0.00 | 0.00 | 0.00 | 0.00 | 0.06 | 0.00 |
| | | 8526 | 0.00 | **1.00** | 0.00 | 0.00 | 0.00 | 0.00 | 0.00 | 0.00 | 0.00 | 0.00 |
| | | 9795 | 0.00 | **1.00** | 0.00 | 0.00 | 0.00 | 0.00 | 0.00 | 0.00 | 0.63 | 0.00 |
| | | 9926 | 0.00 | **1.00** | 0.00 | 0.00 | 0.00 | 0.00 | 0.00 | 0.00 | 0.00 | 0.00 |
| | | 9950 | 0.00 | **1.00** | 0.00 | 0.00 | 0.00 | 0.00 | 0.00 | 0.00 | 0.50 | 0.00 |

Table S5: Spiking Rate of Each Neuron in Algorithmic Model and Device Implementation (Class 2 & Class 3)

| Class | Type | Sample | Neuron | | | | | | | | | |
|---|---|---|---|---|---|---|---|---|---|---|---|---|
| | | | 0 | 1 | 2 | 3 | 4 | 5 | 6 | 7 | 8 | 9 |
| 2 | Algorithm | 0298 | 0.00 | 0.00 | **1.00** | 0.00 | 0.00 | 0.00 | 0.00 | 0.00 | 0.25 | 0.00 |
| | | 0816 | 0.00 | 0.00 | **1.00** | 0.25 | 0.00 | 0.00 | 0.00 | 0.00 | 0.31 | 0.00 |
| | | 1177 | 0.00 | 0.00 | **1.00** | 0.00 | 0.00 | 0.00 | 0.00 | 0.00 | 0.00 | 0.00 |
| | | 2940 | 0.00 | 0.00 | **0.94** | 0.13 | 0.00 | 0.00 | 0.13 | 0.00 | 0.00 | 0.00 |
| | | 2971 | 0.00 | 0.00 | **1.00** | 0.00 | 0.00 | 0.00 | 0.00 | 0.00 | 0.00 | 0.00 |
| | | 3128 | 0.00 | 0.00 | **1.00** | 0.56 | 0.00 | 0.00 | 0.00 | 0.00 | 0.06 | 0.00 |
| | | 4504 | 0.13 | 0.00 | 0.38 | 0.00 | 0.00 | 0.06 | 0.00 | **0.56** | 0.00 | 0.00 |
| | | 7637 | 0.00 | 0.00 | 0.06 | **1.00** | 0.00 | 0.00 | 0.00 | 0.00 | 0.00 | 0.00 |
| | | 7785 | 0.00 | 0.00 | **1.00** | 0.00 | 0.00 | 0.00 | 0.00 | 0.00 | 0.00 | 0.00 |
| | | 9942 | 0.00 | 0.00 | **1.00** | 0.00 | 0.00 | 0.00 | 0.00 | 0.00 | 0.00 | 0.00 |
| | Device | 0298 | 0.00 | 0.00 | **1.00** | 0.00 | 0.00 | 0.00 | 0.00 | 0.00 | 0.25 | 0.00 |
| | | 0816 | 0.00 | 0.00 | **1.00** | 0.25 | 0.00 | 0.00 | 0.00 | 0.00 | 0.44 | 0.00 |
| | | 1177 | 0.00 | 0.00 | **1.00** | 0.00 | 0.00 | 0.00 | 0.00 | 0.00 | 0.00 | 0.00 |
| | | 2940 | 0.00 | 0.00 | **1.00** | 0.19 | 0.00 | 0.00 | 0.31 | 0.00 | 0.13 | 0.00 |
| | | 2971 | 0.00 | 0.00 | **1.00** | 0.00 | 0.00 | 0.00 | 0.00 | 0.00 | 0.00 | 0.00 |
| | | 3128 | 0.00 | 0.00 | **1.00** | 0.81 | 0.00 | 0.00 | 0.00 | 0.00 | 0.13 | 0.00 |
| | | 4504 | 0.06 | 0.00 | **0.69** | 0.00 | 0.00 | 0.06 | 0.00 | 0.50 | 0.00 | 0.00 |
| | | 7637 | 0.00 | 0.00 | 0.13 | **1.00** | 0.00 | 0.00 | 0.00 | 0.00 | 0.00 | 0.00 |
| | | 7785 | 0.00 | 0.00 | **1.00** | 0.00 | 0.00 | 0.00 | 0.00 | 0.00 | 0.00 | 0.00 |
| | | 9942 | 0.00 | 0.00 | **1.00** | 0.00 | 0.00 | 0.00 | 0.00 | 0.00 | 0.00 | 0.00 |
| 3 | Algorithm | 0349 | 0.00 | 0.00 | 0.56 | 0.19 | 0.00 | 0.00 | 0.00 | **0.88** | 0.13 | 0.00 |
| | | 2639 | 0.00 | 0.00 | 0.00 | **1.00** | 0.00 | 0.13 | 0.00 | 0.00 | 0.00 | 0.00 |
| | | 2952 | 0.00 | 0.00 | 0.06 | **1.00** | 0.00 | 0.94 | 0.00 | 0.00 | 0.06 | 0.00 |
| | | 4833 | 0.00 | 0.00 | **1.00** | 1.00 | 0.00 | 0.00 | 0.00 | 0.00 | 0.00 | 0.00 |
| | | 6107 | 0.00 | 0.00 | 0.00 | **1.00** | 0.00 | 0.00 | 0.00 | 0.00 | 0.00 | 0.00 |
| | | 7312 | 0.00 | 0.00 | 0.00 | **1.00** | 0.00 | 0.00 | 0.00 | 0.00 | 0.06 | 0.00 |
| | | 7968 | 0.00 | 0.00 | 0.00 | **1.00** | 0.00 | 0.00 | 0.00 | 0.00 | 0.00 | 0.00 |
| | | 8393 | 0.00 | 0.00 | 0.06 | **1.00** | 0.00 | 0.00 | 0.00 | 0.00 | 0.13 | 0.00 |
| | | 8811 | 0.00 | 0.00 | 0.00 | **1.00** | 0.00 | 0.06 | 0.00 | 0.00 | 0.00 | 0.00 |
| | | 9342 | 0.00 | 0.00 | **1.00** | 0.50 | 0.00 | 0.00 | 0.00 | 0.00 | 0.00 | 0.00 |
| | Device | 0349 | 0.00 | 0.00 | 0.69 | 0.19 | 0.00 | 0.00 | 0.00 | **0.88** | 0.13 | 0.00 |
| | | 2639 | 0.00 | 0.00 | 0.00 | **1.00** | 0.00 | 0.31 | 0.00 | 0.00 | 0.00 | 0.00 |
| | | 2952 | 0.00 | 0.00 | 0.00 | **1.00** | 0.00 | 0.94 | 0.00 | 0.00 | 0.06 | 0.00 |
| | | 4833 | 0.00 | 0.00 | **1.00** | 1.00 | 0.00 | 0.00 | 0.00 | 0.00 | 0.00 | 0.00 |
| | | 6107 | 0.00 | 0.00 | 0.00 | **1.00** | 0.00 | 0.00 | 0.00 | 0.00 | 0.00 | 0.00 |
| | | 7312 | 0.00 | 0.00 | 0.06 | **1.00** | 0.00 | 0.00 | 0.00 | 0.00 | 0.13 | 0.00 |
| | | 7968 | 0.00 | 0.00 | 0.06 | **1.00** | 0.00 | 0.00 | 0.00 | 0.00 | 0.00 | 0.00 |
| | | 8393 | 0.00 | 0.00 | 0.06 | **1.00** | 0.00 | 0.00 | 0.00 | 0.00 | 0.19 | 0.00 |
| | | 8811 | 0.00 | 0.00 | 0.00 | **1.00** | 0.00 | 0.06 | 0.00 | 0.00 | 0.00 | 0.00 |
| | | 9342 | 0.00 | 0.00 | **1.00** | 0.56 | 0.00 | 0.00 | 0.00 | 0.00 | 0.00 | 0.00 |

Table S6: Spiking Rate of Each Neuron in Algorithmic Model and Device Implementation (Class 4 & Class 5)

| Class | Type | Sample | Neuron | | | | | | | | | |
|---|---|---|---|---|---|---|---|---|---|---|---|---|
| | | | 0 | 1 | 2 | 3 | 4 | 5 | 6 | 7 | 8 | 9 |
| 4 | Algorithm | 3718 | 0.00 | 0.00 | 0.00 | 0.00 | 0.88 | 0.00 | 0.00 | 0.00 | 0.00 | **1.00** |
| | | 3792 | 0.00 | 0.00 | 0.00 | 0.00 | **1.00** | 0.00 | 0.00 | 0.00 | 0.00 | 0.00 |
| | | 4266 | 0.00 | 0.00 | 0.00 | 0.00 | **1.00** | 0.00 | 0.00 | 0.00 | 0.00 | 1.00 |
| | | 5926 | 0.00 | 0.00 | 0.00 | 0.00 | **0.94** | 0.00 | 0.00 | 0.00 | 0.00 | 0.88 |
| | | 7341 | 0.00 | 0.00 | 0.00 | 0.00 | **1.00** | 0.00 | 0.00 | 0.06 | 0.00 | 0.63 |
| | | 7456 | 0.00 | 0.00 | 0.00 | 0.00 | **1.00** | 0.00 | 0.00 | 0.00 | 0.00 | 0.50 |
| | | 8193 | 0.00 | 0.00 | 0.00 | 0.00 | **1.00** | 0.00 | 0.00 | 0.00 | 0.00 | 0.63 |
| | | 8312 | 0.00 | 0.00 | 0.00 | 0.00 | **1.00** | 0.00 | 0.00 | 0.00 | 0.00 | 0.13 |
| | | 9099 | 0.00 | 0.00 | 0.00 | 0.00 | **1.00** | 0.00 | 0.00 | 0.00 | 0.00 | 0.00 |
| | | 9605 | 0.00 | 0.00 | 0.00 | 0.00 | **1.00** | 0.00 | 0.00 | 0.00 | 0.00 | 0.06 |
| | Device | 3718 | 0.00 | 0.00 | 0.00 | 0.00 | 0.88 | 0.00 | 0.00 | 0.06 | 0.00 | **1.00** |
| | | 3792 | 0.00 | 0.00 | 0.00 | 0.00 | **1.00** | 0.00 | 0.00 | 0.00 | 0.00 | 0.00 |
| | | 4266 | 0.00 | 0.00 | 0.00 | 0.00 | **1.00** | 0.00 | 0.00 | 0.00 | 0.00 | 1.00 |
| | | 5926 | 0.00 | 0.00 | 0.00 | 0.00 | **0.94** | 0.00 | 0.00 | 0.00 | 0.00 | 0.69 |
| | | 7341 | 0.00 | 0.00 | 0.00 | 0.00 | **1.00** | 0.00 | 0.00 | 0.06 | 0.00 | 0.38 |
| | | 7456 | 0.00 | 0.00 | 0.00 | 0.00 | **1.00** | 0.00 | 0.00 | 0.00 | 0.00 | 0.44 |
| | | 8193 | 0.00 | 0.00 | 0.00 | 0.00 | **1.00** | 0.00 | 0.00 | 0.00 | 0.00 | 0.69 |
| | | 8312 | 0.00 | 0.00 | 0.00 | 0.00 | **1.00** | 0.00 | 0.00 | 0.00 | 0.00 | 0.13 |
| | | 9099 | 0.00 | 0.00 | 0.00 | 0.00 | **1.00** | 0.00 | 0.00 | 0.00 | 0.00 | 0.00 |
| | | 9605 | 0.00 | 0.00 | 0.00 | 0.00 | **1.00** | 0.00 | 0.00 | 0.00 | 0.00 | 0.06 |
| 5 | Algorithm | 0356 | 0.00 | 0.00 | 0.00 | 0.00 | 0.00 | **1.00** | 0.00 | 0.00 | 0.00 | 0.00 |
| | | 1115 | 0.00 | 0.00 | 0.00 | 0.44 | 0.00 | **0.88** | 0.00 | 0.00 | 0.00 | 0.56 |
| | | 1510 | 0.00 | 0.00 | 0.00 | 0.00 | 0.00 | **1.00** | 0.00 | 0.00 | 0.00 | 0.00 |
| | | 1525 | **0.44** | 0.00 | 0.19 | 0.00 | 0.00 | 0.06 | 0.00 | 0.06 | 0.25 | 0.00 |
| | | 2518 | 0.00 | 0.00 | 0.00 | 0.00 | 0.00 | **1.00** | 0.00 | 0.00 | 0.00 | 0.00 |
| | | 2832 | 0.00 | 0.00 | 0.00 | **0.94** | 0.00 | 0.69 | 0.00 | 0.00 | 0.00 | 0.00 |
| | | 4054 | 0.00 | 0.00 | 0.00 | 0.38 | 0.00 | **1.00** | 0.00 | 0.00 | 0.06 | 0.00 |
| | | 4979 | 0.00 | 0.00 | 0.00 | 0.31 | 0.00 | **1.00** | 0.00 | 0.00 | 0.00 | 0.00 |
| | | 5598 | 0.00 | 0.00 | 0.00 | 0.00 | 0.00 | **1.00** | 0.00 | 0.00 | 0.00 | 0.00 |
| | | 7630 | 0.00 | 0.00 | 0.00 | 0.13 | 0.00 | **1.00** | 0.00 | 0.00 | 0.00 | 0.00 |
| | Device | 0356 | 0.00 | 0.00 | 0.00 | 0.00 | 0.00 | **1.00** | 0.00 | 0.00 | 0.00 | 0.00 |
| | | 1115 | 0.00 | 0.00 | 0.00 | 0.56 | 0.00 | **0.88** | 0.00 | 0.00 | 0.00 | 0.50 |
| | | 1510 | 0.00 | 0.00 | 0.00 | 0.06 | 0.00 | **1.00** | 0.00 | 0.00 | 0.00 | 0.00 |
| | | 1525 | **0.44** | 0.00 | 0.13 | 0.06 | 0.00 | 0.13 | 0.00 | 0.00 | 0.31 | 0.00 |
| | | 2518 | 0.00 | 0.00 | 0.00 | 0.00 | 0.00 | **1.00** | 0.00 | 0.00 | 0.00 | 0.00 |
| | | 2832 | 0.00 | 0.00 | 0.00 | **0.94** | 0.00 | 0.63 | 0.00 | 0.00 | 0.06 | 0.00 |
| | | 4054 | 0.00 | 0.00 | 0.00 | 0.38 | 0.00 | **1.00** | 0.00 | 0.00 | 0.06 | 0.00 |
| | | 4979 | 0.00 | 0.00 | 0.00 | 0.50 | 0.00 | **1.00** | 0.00 | 0.00 | 0.00 | 0.00 |
| | | 5598 | 0.00 | 0.00 | 0.00 | 0.00 | 0.00 | **1.00** | 0.00 | 0.00 | 0.00 | 0.00 |
| | | 7630 | 0.00 | 0.00 | 0.00 | 0.31 | 0.00 | **1.00** | 0.00 | 0.00 | 0.00 | 0.00 |

Table S7: Spiking Rate of Each Neuron in Algorithmic Model and Device Implementation (Class 6 & Class 7)

| Class | Type | Sample | Neuron | | | | | | | | | |
|---|---|---|---|---|---|---|---|---|---|---|---|---|
| | | | 0 | 1 | 2 | 3 | 4 | 5 | 6 | 7 | 8 | 9 |
| 6 | Algorithm | 3331 | 0.00 | 0.00 | 0.00 | 0.00 | 0.00 | 0.00 | **1.00** | 0.00 | 0.00 | 0.00 |
| | | 4239 | 0.00 | 0.00 | 0.00 | 0.00 | 0.00 | 0.44 | **0.88** | 0.00 | 0.13 | 0.00 |
| | | 4622 | 0.00 | 0.00 | 0.00 | 0.00 | 0.00 | 0.00 | **1.00** | 0.00 | 0.13 | 0.00 |
| | | 4814 | **0.94** | 0.00 | 0.00 | 0.00 | 0.00 | 0.00 | 0.00 | 0.00 | 0.00 | 0.00 |
| | | 5599 | 0.00 | 0.00 | 0.06 | 0.00 | 0.00 | 0.00 | **1.00** | 0.00 | 0.00 | 0.00 |
| | | 6258 | 0.00 | 0.00 | 0.00 | 0.00 | 0.00 | 0.00 | **1.00** | 0.00 | 0.00 | 0.00 |
| | | 6842 | 0.00 | 0.00 | 0.13 | 0.00 | 0.00 | 0.00 | **1.00** | 0.00 | 0.00 | 0.00 |
| | | 8423 | 0.00 | 0.00 | 0.13 | 0.00 | 0.00 | 0.00 | **1.00** | 0.00 | 0.00 | 0.00 |
| | | 8990 | 0.00 | 0.00 | 0.00 | 0.00 | 0.00 | 0.00 | **1.00** | 0.00 | 0.00 | 0.00 |
| | | 9149 | 0.00 | 0.00 | 0.00 | 0.00 | 0.00 | 0.00 | **1.00** | 0.00 | 0.00 | 0.00 |
| | Device | 3331 | 0.00 | 0.00 | 0.00 | 0.00 | 0.00 | 0.00 | **1.00** | 0.00 | 0.00 | 0.00 |
| | | 4239 | 0.06 | 0.00 | 0.00 | 0.00 | 0.00 | 0.44 | **0.94** | 0.00 | 0.13 | 0.00 |
| | | 4622 | 0.00 | 0.00 | 0.00 | 0.00 | 0.00 | 0.00 | **1.00** | 0.00 | 0.13 | 0.00 |
| | | 4814 | **0.94** | 0.00 | 0.00 | 0.00 | 0.00 | 0.00 | 0.00 | 0.00 | 0.00 | 0.00 |
| | | 5599 | 0.00 | 0.00 | 0.06 | 0.00 | 0.00 | 0.00 | **1.00** | 0.00 | 0.00 | 0.00 |
| | | 6258 | 0.00 | 0.00 | 0.00 | 0.00 | 0.00 | 0.00 | **1.00** | 0.00 | 0.00 | 0.00 |
| | | 6842 | 0.00 | 0.00 | 0.13 | 0.00 | 0.00 | 0.00 | **1.00** | 0.00 | 0.00 | 0.00 |
| | | 8423 | 0.00 | 0.00 | 0.13 | 0.00 | 0.00 | 0.00 | **1.00** | 0.00 | 0.00 | 0.06 |
| | | 8990 | 0.00 | 0.00 | 0.00 | 0.00 | 0.00 | 0.00 | **1.00** | 0.00 | 0.00 | 0.00 |
| | | 9149 | 0.00 | 0.00 | 0.00 | 0.00 | 0.00 | 0.00 | **1.00** | 0.00 | 0.00 | 0.00 |
| 7 | Algorithm | 0617 | 0.00 | 0.00 | 0.63 | 0.00 | 0.00 | 0.00 | 0.00 | **1.00** | 0.00 | 0.00 |
| | | 2091 | 0.00 | 0.00 | 0.00 | 0.00 | 0.00 | 0.00 | 0.00 | **1.00** | 0.00 | 0.00 |
| | | 3225 | 0.00 | 0.56 | 0.19 | 0.44 | 0.00 | 0.06 | 0.00 | **0.63** | 0.13 | 0.00 |
| | | 4214 | 0.00 | 0.00 | 0.00 | 0.00 | 0.13 | 0.00 | 0.00 | **1.00** | 0.00 | 0.13 |
| | | 4693 | 0.00 | 0.00 | 0.00 | 0.00 | 0.00 | 0.00 | 0.00 | **1.00** | 0.00 | 0.44 |
| | | 5600 | 0.00 | 0.00 | 0.00 | 0.00 | 0.06 | 0.00 | 0.00 | 0.75 | 0.19 | **1.00** |
| | | 6589 | 0.00 | 0.00 | 0.31 | 0.00 | 0.00 | 0.00 | 0.00 | **1.00** | 0.00 | 0.00 |
| | | 6762 | 0.00 | 0.00 | 0.88 | 0.00 | 0.00 | 0.00 | 0.00 | **1.00** | 0.00 | 0.00 |
| | | 7069 | 0.00 | 0.00 | 0.00 | 0.00 | 0.00 | 0.00 | 0.00 | **1.00** | 0.00 | 0.00 |
| | | 8248 | 0.00 | 0.00 | 0.00 | 0.00 | 0.00 | 0.00 | 0.00 | **1.00** | 0.00 | 0.00 |
| | Device | 0617 | 0.00 | 0.00 | 0.63 | 0.00 | 0.00 | 0.00 | 0.00 | **1.00** | 0.00 | 0.00 |
| | | 2091 | 0.00 | 0.00 | 0.00 | 0.00 | 0.00 | 0.00 | 0.00 | **1.00** | 0.00 | 0.00 |
| | | 3225 | 0.00 | 0.50 | 0.25 | 0.50 | 0.00 | 0.00 | 0.00 | **0.63** | 0.13 | 0.06 |
| | | 4214 | 0.00 | 0.00 | 0.00 | 0.00 | 0.06 | 0.00 | 0.00 | **1.00** | 0.00 | 0.13 |
| | | 4693 | 0.00 | 0.00 | 0.00 | 0.00 | 0.00 | 0.00 | 0.00 | **1.00** | 0.00 | 0.56 |
| | | 5600 | 0.00 | 0.00 | 0.00 | 0.00 | 0.00 | 0.00 | 0.00 | 0.75 | 0.31 | **1.00** |
| | | 6589 | 0.00 | 0.00 | 0.38 | 0.00 | 0.00 | 0.00 | 0.00 | **1.00** | 0.00 | 0.00 |
| | | 6762 | 0.00 | 0.00 | 0.81 | 0.00 | 0.00 | 0.00 | 0.00 | **1.00** | 0.00 | 0.00 |
| | | 7069 | 0.00 | 0.00 | 0.00 | 0.00 | 0.00 | 0.00 | 0.00 | **1.00** | 0.00 | 0.00 |
| | | 8248 | 0.00 | 0.00 | 0.00 | 0.00 | 0.00 | 0.00 | 0.00 | **1.00** | 0.00 | 0.00 |

Table S8: Spiking Rate of Each Neuron in Algorithmic Model and Device Implementation (Class 8 & Class 9)

| Class | Type | Sample | Neuron | | | | | | | | | |
|---|---|---|---|---|---|---|---|---|---|---|---|---|
| | | | 0 | 1 | 2 | 3 | 4 | 5 | 6 | 7 | 8 | 9 |
| 8 | Algorithm | 0714 | 0.06 | 0.00 | 0.00 | 0.13 | 0.00 | 0.25 | 0.00 | 0.00 | **1.00** | 0.00 |
| | | 1371 | 0.00 | 0.00 | 0.00 | 0.31 | 0.00 | 0.00 | 0.00 | 0.00 | **1.00** | 0.00 |
| | | 2859 | 0.00 | 0.06 | 0.19 | 0.00 | 0.00 | 0.00 | 0.19 | 0.00 | **1.00** | 0.00 |
| | | 3064 | 0.00 | 0.00 | 0.00 | 0.31 | 0.00 | 0.00 | 0.00 | 0.00 | **1.00** | 0.00 |
| | | 5049 | 0.00 | 0.00 | 0.00 | 0.00 | 0.00 | 0.00 | 0.13 | 0.00 | **0.94** | 0.00 |
| | | 6001 | 0.00 | 0.00 | 0.00 | 0.06 | 0.00 | 0.00 | 0.00 | 0.00 | **1.00** | 0.00 |
| | | 6617 | 0.00 | 0.00 | 0.06 | 0.06 | 0.00 | 0.00 | 0.00 | 0.00 | **1.00** | 0.25 |
| | | 6654 | 0.00 | 0.00 | 0.00 | 0.00 | 0.00 | 0.00 | 0.00 | 0.00 | **1.00** | 0.00 |
| | | 8699 | 0.00 | 0.00 | 0.00 | 0.00 | 0.00 | 0.00 | 0.00 | 0.00 | **1.00** | 0.00 |
| | | 8934 | 0.00 | 0.00 | 0.00 | 0.00 | 0.00 | 0.00 | 0.00 | 0.00 | **1.00** | 0.00 |
| | Device | 0714 | 0.06 | 0.00 | 0.00 | 0.13 | 0.00 | 0.25 | 0.00 | 0.00 | **1.00** | 0.00 |
| | | 1371 | 0.00 | 0.00 | 0.00 | 0.44 | 0.00 | 0.00 | 0.00 | 0.00 | **1.00** | 0.00 |
| | | 2859 | 0.00 | 0.00 | 0.19 | 0.00 | 0.00 | 0.00 | 0.31 | 0.00 | **1.00** | 0.00 |
| | | 3064 | 0.00 | 0.00 | 0.00 | 0.31 | 0.00 | 0.00 | 0.00 | 0.00 | **1.00** | 0.00 |
| | | 5049 | 0.00 | 0.00 | 0.00 | 0.00 | 0.00 | 0.00 | 0.19 | 0.00 | **1.00** | 0.00 |
| | | 6001 | 0.00 | 0.00 | 0.00 | 0.06 | 0.00 | 0.00 | 0.00 | 0.00 | **1.00** | 0.00 |
| | | 6617 | 0.00 | 0.00 | 0.13 | 0.06 | 0.00 | 0.00 | 0.00 | 0.00 | **1.00** | 0.25 |
| | | 6654 | 0.00 | 0.00 | 0.00 | 0.00 | 0.00 | 0.00 | 0.00 | 0.00 | **1.00** | 0.00 |
| | | 8699 | 0.00 | 0.00 | 0.00 | 0.00 | 0.00 | 0.00 | 0.00 | 0.00 | **1.00** | 0.00 |
| | | 8934 | 0.00 | 0.00 | 0.00 | 0.00 | 0.00 | 0.00 | 0.00 | 0.00 | **1.00** | 0.00 |
| 9 | Algorithm | 0193 | 0.00 | 0.00 | 0.00 | **0.38** | 0.25 | 0.13 | 0.00 | 0.13 | 0.00 | 0.13 |
| | | 0639 | 0.00 | 0.00 | 0.00 | 0.00 | 0.06 | 0.00 | 0.00 | 0.00 | 0.00 | **1.00** |
| | | 0962 | 0.00 | 0.00 | 0.00 | 0.00 | 0.31 | 0.00 | 0.00 | 0.94 | 0.31 | **1.00** |
| | | 1597 | 0.00 | 0.00 | 0.00 | 0.00 | 0.38 | 0.00 | 0.00 | 0.19 | 0.06 | **1.00** |
| | | 3041 | 0.00 | 0.00 | 0.00 | 0.00 | 0.25 | 0.00 | 0.00 | 0.13 | 0.06 | **1.00** |
| | | 3723 | 0.00 | 0.00 | 0.13 | 0.00 | 0.06 | 0.00 | 0.00 | 0.00 | 0.06 | **1.00** |
| | | 4237 | 0.00 | 0.00 | 0.00 | 0.00 | 0.06 | 0.00 | 0.00 | 0.00 | 0.00 | **1.00** |
| | | 4761 | 0.00 | 0.00 | 0.06 | 0.19 | 0.13 | 0.19 | 0.00 | 0.00 | **0.50** | 0.06 |
| | | 8002 | 0.00 | 0.00 | 0.00 | 0.00 | 0.00 | 0.00 | 0.00 | 0.56 | 0.00 | **1.00** |
| | | 8998 | 0.00 | 0.00 | 0.00 | 0.00 | 0.00 | 0.00 | 0.00 | 0.44 | 0.00 | **1.00** |
| | Device | 0193 | 0.00 | 0.00 | 0.06 | **0.50** | 0.31 | 0.13 | 0.00 | 0.13 | 0.06 | 0.13 |
| | | 0639 | 0.00 | 0.00 | 0.00 | 0.00 | 0.06 | 0.00 | 0.00 | 0.00 | 0.00 | **1.00** |
| | | 0962 | 0.00 | 0.00 | 0.00 | 0.00 | 0.25 | 0.00 | 0.00 | 0.88 | 0.44 | **1.00** |
| | | 1597 | 0.00 | 0.00 | 0.00 | 0.00 | 0.44 | 0.00 | 0.00 | 0.25 | 0.06 | **1.00** |
| | | 3041 | 0.00 | 0.00 | 0.00 | 0.00 | 0.19 | 0.00 | 0.00 | 0.25 | 0.06 | **1.00** |
| | | 3723 | 0.00 | 0.00 | 0.19 | 0.00 | 0.00 | 0.00 | 0.00 | 0.00 | 0.06 | **1.00** |
| | | 4237 | 0.00 | 0.00 | 0.00 | 0.00 | 0.06 | 0.00 | 0.00 | 0.00 | 0.00 | **1.00** |
| | | 4761 | 0.00 | 0.00 | 0.06 | 0.38 | 0.19 | 0.13 | 0.00 | 0.00 | **0.75** | 0.06 |
| | | 8002 | 0.00 | 0.00 | 0.00 | 0.00 | 0.00 | 0.00 | 0.00 | 0.56 | 0.00 | **1.00** |
| | | 8998 | 0.00 | 0.00 | 0.00 | 0.00 | 0.06 | 0.00 | 0.00 | 0.44 | 0.00 | **1.00** |

# REFERENCES AND NOTES.